\documentclass[11pt]{article}

\PassOptionsToPackage{numbers, compress}{natbib}
\usepackage[utf8]{inputenc}

\usepackage{amsmath}
\DeclareMathOperator*{\argmax}{arg\,max}
\DeclareMathOperator*{\argmin}{arg\,min}
\usepackage{amsfonts}
\usepackage{mathrsfs}
\usepackage{amssymb}
\usepackage{amsthm}

\usepackage{libertine}
\usepackage{wrapfig}
\usepackage{graphicx}
\usepackage{multirow}
\usepackage{epstopdf}
\usepackage{multicol}
\usepackage{subfigure}
\usepackage{booktabs}

\usepackage{algorithmic}
\usepackage{algorithm}
\usepackage[algo2e,linesnumbered]{algorithm2e}
\usepackage{stackengine}
\usepackage{graphicx}
\usepackage{multirow}
\usepackage{epstopdf}
\usepackage{multicol}
\usepackage{caption}
\usepackage{subfigure}
\usepackage{enumitem}
\usepackage{amsmath}
\usepackage{pifont}
\usepackage{natbib}
\usepackage{bbm}
\usepackage{makecell}
\usepackage{mathtools}
\usepackage[table]{xcolor}
\definecolor{mydarkred}{rgb}{0.6,0,0}
\definecolor{mydarkgreen}{rgb}{0,0.6,0}
\usepackage[colorlinks,
linkcolor=mydarkred,
citecolor=mydarkgreen]{hyperref}
\usepackage{url}
\usepackage{bm}
\usepackage{pifont}
\usepackage{stfloats}
\usepackage[T1]{fontenc}
\usepackage{arydshln}
\usepackage{colortbl}
\usepackage{balance}

\usepackage{wrapfig}
\usepackage{multirow}
\usepackage{multicol}

\newtheorem{theorem}{Theorem}
\newtheorem{assumption}{Assumption}
\newtheorem{definition}{Definition}[section]

\newcolumntype{L}[1]{>{\raggedright\let\newline\\\arraybackslash\hspace{0pt}}m{#1}}
\newcolumntype{Y}{>{\centering\arraybackslash}X}
\newcolumntype{s}{>{\hsize=.3\hsize}Y}
\newcolumntype{t}{>{\hsize=1.5\hsize}X}
\newcolumntype{u}{>{\hsize=0.8\hsize}Y}
\usepackage{makecell}
\usepackage{mathtools}
\usepackage[utf8]{inputenc}
\usepackage{stfloats}

\title{Making Binary Classification from Multiple Unlabeled Datasets Almost Free of Supervision}
       
\author{
Yuhao Wu$^{1}$, Xiaobo Xia$^{1}$, Jun Yu$^{2}$, Bo Han$^{3}$,\\ Gang Niu$^{4}$, Masashi Sugiyama$^{4,5}$, Tongliang Liu$^{1}$ \\
  \small{$^1$Sydney AI Centre, The University of Sydney;}\\
  \small{$^2$Department of Automation, University of Science and Technology of China;}\\
  \small{$^3$Department of Computer Science, Hong Kong Baptist University;}\\
  \small{$^4$Center for Advanced Intelligence Project, RIKEN;}\\
  \small{$^5$Graduate School of Frontier Sciences, The University of Tokyo}
}
\date{}
\begin{document}

\maketitle

\begin{abstract}

Training a classifier exploiting a huge amount of supervised data is expensive or even prohibited in a situation, where the labeling cost is high. The remarkable progress in working with weaker forms of supervision is binary classification from multiple unlabeled datasets which requires the knowledge of exact class priors for all unlabeled datasets. However, the availability of class priors is restrictive in many real-world scenarios. To address this issue, we propose to solve a new problem setting, i.e., binary classification from \underline{m}ultiple \underline{u}nlabeled datasets with only \underline{o}ne \underline{p}airwise numerical relationshi\underline{p} of class pri\underline{o}rs (MU-OPPO), which knows the relative order (which unlabeled dataset has a higher proportion of positive examples) of two class-prior probabilities for two datasets among multiple unlabeled datasets. In MU-OPPO, we do not need the class priors for all unlabeled datasets, but we only require that there exists a pair of unlabeled datasets for which we know which unlabeled dataset has a larger class prior. Clearly, this form of supervision is easier to be obtained, which can make labeling costs almost free. We propose a novel framework to handle the MU-OPPO problem, which consists of four sequential modules: (i) pseudo label assignment; (ii) confident example collection; (iii) class prior estimation; (iv) classifier training with estimated class priors. Theoretically, we analyze the gap between estimated class priors and true class priors under the proposed framework. Empirically, we confirm the superiority of our framework with comprehensive experiments. Experimental results demonstrate that our framework brings smaller estimation errors of class priors and better performance of binary classification. 

\end{abstract}
\vspace{1em}
\newpage

\section{Introduction} \label{sec:1}

Deep learning with large-scale supervised data has enjoyed huge successes in various domains~\cite{goodfellow2016deep,boecking2020interactive,mazzetto2021adversarial,northcutt2021pervasive,yang2022objects}. However, it is often very costly or even infeasible to collect the data with strong supervision~\cite{scott2015rate,ICLR:Lu+etal:2019,ruhling2021end,bach2017learning}. The fact motivates us to investigate learning algorithms that work with weaker forms of supervision~\cite{lu2021binary,sugiyama2022machine,awasthi2020learning,gu2022instance,varma2019learning}, e.g., partial labels~\cite{xie2018partial,wang2022pico,zhang2022exploiting,cour2011learning} and noisy labels~\cite{lukasik2020does,menon2019can,xia2020part,harutyunyan2020improving,Wei_2020_CVPR,wei2023logitclipping}.

The remarkable progress in working with weaker forms of supervision is \textit{binary classification from multiple unlabeled datasets}~\cite{scott2020learning,lu2021binary,ICLR:Lu+etal:2019,du2013clustering}. Those multiple unlabeled datasets share the same class-conditional density and the aim is to learn a \textit{binary classifier} from $m$ $(m\geq 2)$ unlabeled datasets with different class priors, i.e., the proportion of positives in each unlabeled dataset~\cite{lu2021binary}. Such a learning scheme is credible in practice. Unlabeled datasets with different class priors can be naturally collected. For example, considering morbidity rates, they can be potential patient data collected from different regions. The rates are likely to be very different because of the different region backgrounds \cite{croft2018urban}.

Class priors play essential roles in the problem of binary classification from multiple unlabeled datasets. In detail, class priors can be leveraged as the supervision to make the problem mathematically solvable, further leading to \textit{statistically consistent} methods~\cite{lu2021binary}. Prior works assumed that the class priors of multiple unlabeled datasets are given, which is reflected in Table~\ref{tab:motivation}. However, in many real-world scenarios, the class priors are \textit{unavailable}~\cite{menon2015learning}. In addition, randomly set class priors or poorly estimated class priors cannot work well (see empirical evidence in Section~\ref{sec:2.2}). It is still mysterious nowadays that the solution for binary classification from multiple unlabeled datasets \textit{without} given class priors.

\begin{figure*}[!t] 
    \centering
    \includegraphics[width=\linewidth]{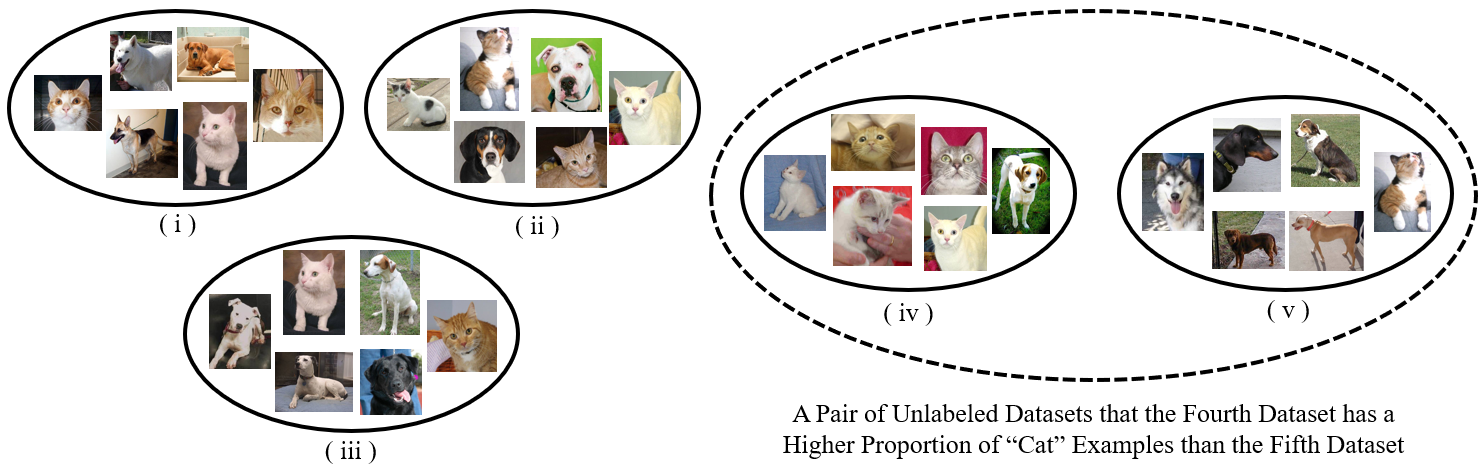}
    \begin{center}
  \caption{An illustration of the MU-OPPO problem setting. We have five unlabeled datasets and regard the images of cats (resp. dogs) as positive (resp. negative) examples. In this MU-OPPO problem, there are no given class priors (the proportions of cats in the datasets) for five unlabeled datasets. The only requirement is that there exists a dataset pair (dataset (iv) and dataset (v)) where we know dataset (iv) has a larger class prior (the proportion of cats in the dataset) than dataset (v).}
    \label{illustration_setting}
    \end{center}
    \vspace{-2em}
\end{figure*}

In this paper, we raise a new problem setting, i.e., \underline{m}ultiple \underline{u}nlabeled datasets with only \underline{o}ne \underline{p}airwisely numerical relationshi\underline{p} of class pri\underline{o}rs (MU-OPPO). In this new problem illustrated in Figure~\ref{illustration_setting}, there are no given class priors for multiple unlabeled datasets that share the same class-conditional density. Instead, multiple unlabeled datasets are provided with at least one pairwise \textit{numerical relationship} of class priors. That is to say, we only require that there exists a pair of unlabeled datasets for which we know which unlabeled dataset has a larger class prior. This weakly supervised requirement makes binary classification from multiple unlabeled
datasets almost free of supervision.  
For example, considering the data about morbidity rates of patients in different regions, we cannot obtain precise morbidity rates directly without careful diagnoses, which could be very costly. On the contrary, we can easily obtain that, for a pair of regions, which one has a higher morbidity rate than the other according to relevant and easily available knowledge, e.g., the region having poor sanitary and medical conditions usually has a high morbidity rate. Generally speaking, obtaining the numerical relationship of two class priors would be much easier than directly obtaining precise class priors and thus less costly.  Although the supervision in MU-OPPO is more accessible by comparing precise class priors with the pairwise numerical relationship of class priors, the learning task becomes more challenging since the supervised information of the pairwise numerical relationship of class priors is less than exact class priors.

To address the MU-OPPO problem, we propose a novel learning framework in this paper. Specifically, our learning framework consists of four modules: (i) pseudo label assignment; (ii) confident example collection; (iii) class prior estimation; (iv) classifier training with estimated class priors. We handle the MU-OPPO problem by executing the four modules sequentially. The illustration of our proposed framework is provided in Figure~\ref{illustration}. In particular, we first assign pseudo labels to the two unlabeled datasets whose class prior numerical relationship is known. Specifically, the unlabeled dataset having a larger (resp. smaller) positive class prior is assigned with positive (resp. negative) pseudo labels. Then, we select confident data whose pseudo labels are more likely to be correct, from the two datasets with pseudo labels. Afterward, with the selected confident data, the class priors of all unlabeled datasets can be estimated under irreducible and mutually irreducible assumptions~\cite{scott2015rate}. In the end, the estimated class priors are then employed to build statistically consistent algorithms for binary classification from unlabeled datasets~\cite{ICLR:Lu+etal:2019,lu2021binary}.

\paragraph{Contributions.} Before delving into details, we summarize the contributions of this paper as follows. 

\begin{table}[!t] \footnotesize
    \centering
    \caption{Comparison of previous methods designed for binary classification from unlabeled datasets.}
    \begin{tabular}{l|ccc}
    \hline
         \makecell[c]{Methods} & \makecell[c]{Deal with \\ Two Sets} & \makecell[c]{Deal with \\ Three or More Sets} & \makecell[c]{Unknown \\Class Priors}\\

         \hline
         XR~\cite{mann2007simple} & $\checkmark$ & $\times$ & $\times$ \\
         InvCal~\cite{rueping2010svm} & $\checkmark$ & $\checkmark$ & $\times$ \\
         LSDD~\cite{du2013clustering} & $\checkmark$ & $\times$ & $\times$ \\
         Proportion-SVM~\cite{felix2013psvm} & $\checkmark$ & $\checkmark$ & $\times$ \\
         BER~\cite{menon2015learning} & $\checkmark$ & $\times$ & $\times$ \\
         UU~\cite{ICLR:Lu+etal:2019}& $\checkmark$ & $\times$ & $\times$ \\
         UU-c~\cite{lu2020mitigating}& $\checkmark$ & $\times$ & $\times$ \\
         LLP-VAT~\cite{tsai2020learning} & $\checkmark$ & $\checkmark$ & $\times$ \\
         MCM~\cite{scott2020learning} & $\checkmark$ & $\checkmark$ & $\times$ \\
         U$^m$-SSC~\cite{lu2021binary} & $\checkmark$ & $\checkmark$ & $\times$\\
        
         \hline
         \textbf{Ours} & $\checkmark$ & $\checkmark$ & $\checkmark$\\
         \hline
    \end{tabular}

    \label{tab:motivation}
\end{table}

\begin{itemize}
    \item We propose a new and realistic problem setting that targets binary classification from multiple unlabeled datasets with only one pairwise numerical relationship of class priors (MU-OPPO). Unlike previous proposals (see Table~\ref{tab:motivation}), the proposed problem setting gets rid of given class priors, which requires almost free supervision information~(Section \ref{sec:2.1}).
    
    \item We establish a generalized framework for the MU-OPPO problem, which contains four modules, as discussed above. Within the framework, multiple class priors can be estimated well by utilizing only one pairwise numerical relationship of class priors. This makes applying existing statistically consistent binary classification algorithms from unlabeled datasets successful~(Section \ref{sec:3}).
    
    \item Theoretically, we analyze the gap between estimated class priors and true class priors under the proposed framework.~(Section \ref{sec:5}) Empirically, we evaluate the effectiveness of the proposed framework with comprehensive experiments based on the problem setting of MU-OPPO. Besides, we verify the generalization capability of our framework by replacing various methods in different modules. Experimental results confirm the advancement of the framework~(Section \ref{sec:6}).

\end{itemize}

\paragraph{Organization.} The rest of this paper is organized as follows. In Section~\ref{sec:2}, we formally set up the MU-OPPO problem and analyze the dilemma when previous methods target this problem. In Section~\ref{sec:3}, we discuss the proposed framework and include modules in detail. In Section~\ref{sec:4},  we discuss the difference between the proposed framework and the previous advanced one. In Section~\ref{sec:5}, the theoretical analysis provides the gap between estimated class priors and true class priors under the proposed framework. In Section~\ref{sec:6}, a series of experiments are presented to verify the effectiveness of the proposed framework. In Section~\ref{sec:7}, we provide a more comprehensive review of related literature. Lastly, in Section~\ref{sec:8}, we conclude this paper. 

\section{Preliminaries}\label{sec:2}
In this section, we first formulate the MU-OPPO problem (Section~\ref{sec:2.1}). Then, we show that prior solutions cannot be directly applied to solve the MU-OPPO problem, which demonstrates the necessity of the proposed learning paradigm (Section~\ref{sec:2.2}). 
\subsection{Problem Statement}\label{sec:2.1}
We warm up with binary classification in supervised learning. Let $\mathcal{X}$ be the input feature space and $\mathcal{Y}=\{+1,-1\}$ be the binary label space respectively. Let $\bm{x}\in\mathcal{X}$ and $y\in\mathcal{Y}$ denote the input and output random variables, following an underlying joint distribution $\mathcal{D}$. In supervised learning, the goal of binary classification is to train a classifier $f:\mathcal{X}\rightarrow\mathcal{Y}$ 
that minimizes the risk defined as
\begin{equation}
    R(f):=\mathbb{E}_{(\bm{x},y)\sim\mathcal{D}}[\ell(f(\boldsymbol{x}),y)],
\end{equation}
where $\mathbb{E}$ denotes the expectation and $\ell:\mathbbm{R}\times\mathcal{Y}\rightarrow[0,+\infty)$ is the specific loss function, e.g., the cross-entropy loss~\cite{mohri2018foundations}. In most cases, $R(f)$ cannot be calculated directly since  the joint distribution $\mathcal{D}$ is unknown to the learner. Instead, we are given a fully labeled training dataset $\{(\bm{x}_i,y_i)\}_{i=1}^n$ i.i.d.~drawn from $\mathcal{D}$, where $n$ is the size of training examples. The empirical risk is then used to approximate $R(f)$ by 
\begin{equation}
    \hat{R}(f):=\frac{1}{n}\sum_{i=1}^n\ell(f(\bm{x}_i),y_i).
\end{equation}
Compared with binary classification in supervised learning, MU-OPPO considers a different problem set.

\paragraph{Problem Set.} In MU-OPPO, we only have access to $m$ $(m\geq 2)$ unlabeled datasets denoted by $\mathcal{X}_{\mathrm{u}}=\{\mathcal{X}_{\mathrm{u}}^{j}\}_{j=1}^{m}$. Here, $\mathcal{X}_{\mathrm{u}}^{j} = \{\bm{x}_{1}^j,\ldots, \boldsymbol{x}_{\mathrm{n}_j}^j\} \overset{\rm{i.i.d.}}{\sim} \mathbbm{P}_{\mathrm{u}}^{j}(\bm{x})$, where $n_j$ and $\mathbbm{P}_{\mathrm{u}}^{j}(\boldsymbol{x})$ denote the sample size and the marginal density of the $j$-th unlabeled dataset respectively. The marginal density can be seen as a mixture of the positive and negative class-conditional density $[\mathbbm{P}_{\mathrm{p}}(\bm{x}),\mathbbm{P}_{\mathrm{n}}(\bm{x})]:=[\mathbbm{P}(\bm{x}|y=+1),\mathbbm{P}(\bm{x}|y=-1)]$ by the class priors $\pi_j=\mathbbm{P}_{\mathrm{u}}^j(y=+1)$, i.e., 
\begin{equation}
\mathbbm{P}_\mathrm{u}^j(\bm{x})=\pi_j \mathbbm{P}_\mathrm{p}(\bm{x}) + (1-\pi_j)\mathbbm{P}_\mathrm{n}(\bm{x}).
\label{ptr}
\end{equation}
For all $m$ unlabeled datasets, there exists one pair of unlabeled datasets, e.g., $\mathcal{X}_\mathrm{u}^{\alpha}$ and $\mathcal{X}_\mathrm{u}^{\beta}$, the numerical relationship of their corresponding class priors $\pi_{\alpha}$ and $\pi_{\beta}$ are known as weak supervision, i.e., $\pi_{\alpha}>\pi_{\beta}$ or $\pi_{\alpha}<\pi_{\beta}$\footnote{According to \cite{scott2015rate} and \cite{lu2021binary}, among the $m$ unlabeled datasets, to make the problem mathematically solvable, it is necessary to assume that at least two of class priors $\{\mathcal{\pi}_{j}\}_{j=1}^{m}$ are different, i.e., $\exists~j,j'\in\{1,\ldots,m\}$ such that $j \neq j'$ and $\pi_j \neq \pi_{j'}$.}.

Compared with our problem setting, prior works~\cite{ICLR:Lu+etal:2019,lu2021binary,scott2020learning,tsai2020learning} assumed that the class priors $\pi_j$ for any $j\in\{1,\ldots,m\}$ is known. As discussed in Section \ref{sec:1}, in many realistic scenarios, the class priors are unavailable. Therefore, we have introduced the problem of MU-OPPO above. Obviously, the supervision of one pairwise numerical relationship of class priors is weaker than multiple exact class priors. If the exact class priors are available, then the pairwise numerical relationship of class priors can be accessed, but not vice versa. Although the supervision signal in our problem setting is almost free, the goal is still to obtain
a binary classifier that can generalize well with respect to the joint distribution $\mathcal{D}$, even though it is unobserved.

\begin{figure*}[!t] 
    \centering
    \includegraphics[width=\linewidth,height=0.5\linewidth]{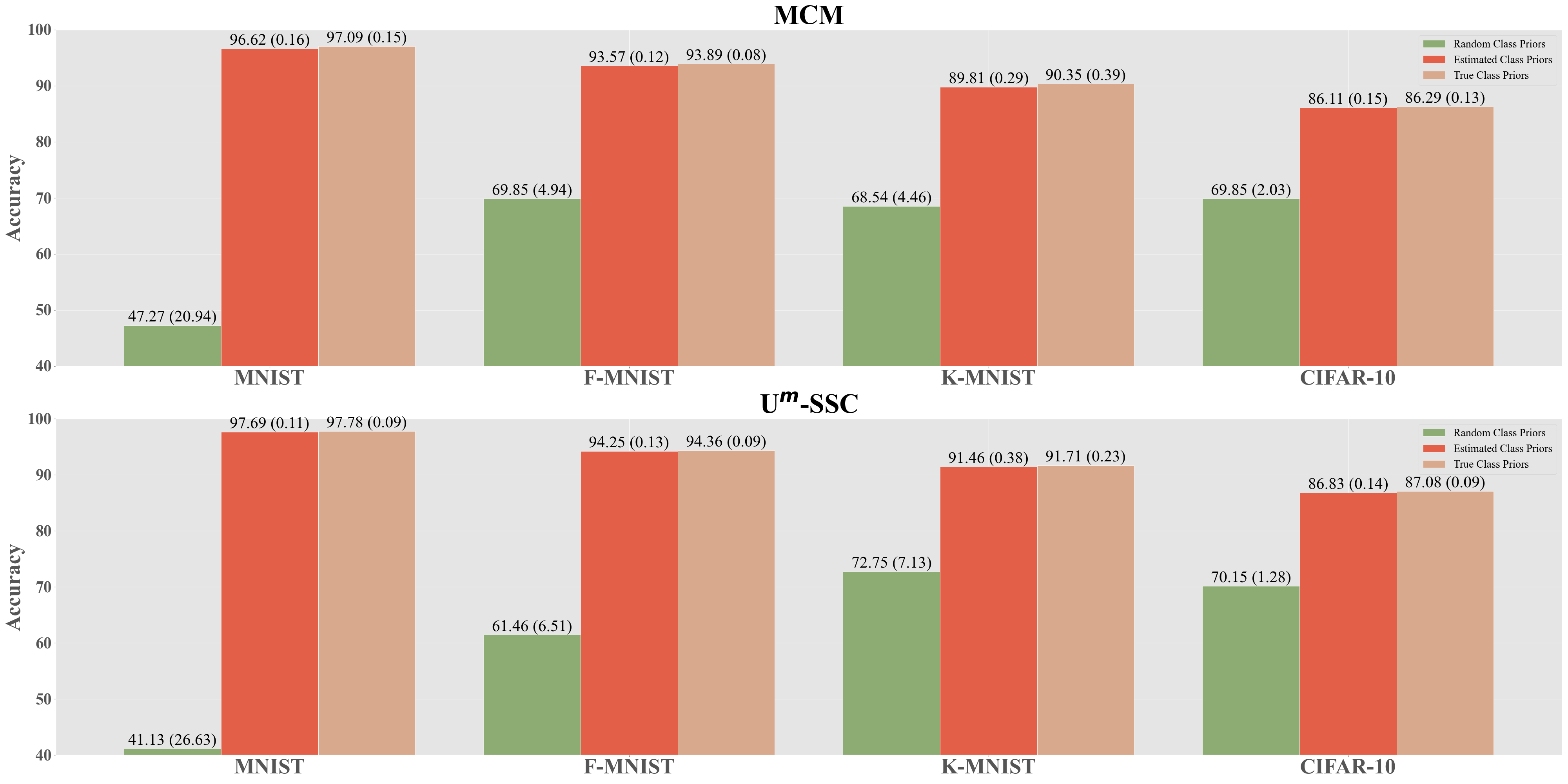}
    \begin{center}

  \caption{There are two state-of-the-art frameworks for binary classification from multiple unlabeled datasets, each with different types of class priors. Specifically, random class priors are generated from the truncated $\mathcal{N}(0.5,1)$ within the interval $(0,1)$, while estimated class priors are estimated from our proposed framework. Note that the methods \underline{m}utual \underline{c}ontamination \underline{m}odels (MCM) ~\cite{scott2020learning} and U$^m$-\underline{s}urrogate \underline{s}et \underline{c}lassification (SSC)~\cite{lu2021binary} trained models on four public datasets. The corresponding classification accuracy and standard deviation are reported at the top of the bins. }
    \label{important}
    \end{center}
\vspace{-2em}
\end{figure*}

\subsection{The Challenging of MU-OPPO}\label{sec:2.2}
 
In MU-OPPO, exact class priors of multiple unlabeled datasets are unavailable, which makes the problem challenging. For estimating class priors, the state-of-the-art methods are highly related to \underline{m}ixture \underline{p}roportion \underline{e}stimation (MPE)~\cite{ramaswamy2016mixture,ivanov2020dedpul,yao2022rethinking}. However, those methods cannot address our MU-OPPO problem, and we will explain the reasons below.

In the setting of MPE, there is a mixture distribution $\mathcal{F}$, i.e., 
\begin{equation}
    \mathcal{F}=(1-\kappa^{*})\mathcal{G}+\kappa^{*}\mathcal{H},
\end{equation}
where $\mathcal{G}$ and $\mathcal{H}$ denote two component distributions. Given the examples randomly collected from $\mathcal{F}$ and $\mathcal{H}$, MPE aims to estimate the maximum proportion $\kappa^{*} \in (0,1)$ of $\mathcal{H}$ in $\mathcal{F}$~\cite{yao2022rethinking,garg2021mixture}. Ideally, if we regard $\kappa^{*}$ as the class prior, the MPE methods can be applied to estimate $\pi_j$ by replacing $\mathcal{F}$ and $\mathcal{H}$ with the mixture distribution $\mathbbm{P}_{\mathrm{u}}^{j}(\bm{x})$ and one of the component distribution $\mathbbm{P}_\mathrm{p}^j(\bm{x})$. 

Unfortunately, in MU-OPPO, this direct application is infeasible. The reason is that $\mathbbm{P}_\mathrm{p}$ is latent, while only $\mathbbm{P}_\mathrm{u}$ is obtainable. Moreover, Scott (2015) applied a mutual MPE model (Scott, 2015) to the setting of learning from noisy labels with the rigorous assumption of class priors. Although their method is related to our work, we show in Section~\ref{sec:4} that this method has drawbacks when applied to our MU-OPPO setting. 
Note that randomly generating class priors cannot perform well for binary classification, which is verified in Figure~\ref{important}. 
Therefore, to accurately estimate class priors, we propose a novel framework in Section \ref{sec:3}, where it is valid to approximate the latent distributions $\mathbbm{P}_\mathrm{p}$ and $\mathbbm{P}_\mathrm{n}$ out of mixture distributions $\mathbbm{P}_{\mathrm{u}}^{j}(\bm{x})$ with theoretical guarantees. We call the proposed framework for estimating the class priors the \underline{c}onfident \underline{c}lass-\underline{p}rior \underline{e}stimator (CCPE) since we use a collection of confident examples in approximating the latent distributions $\mathbbm{P}_\mathrm{p}$ and $\mathbbm{P}_\mathrm{n}$.

\section{Tackling the MU-OPPO Problem with a Solution Framework}\label{sec:3} 

\begin{figure*}[!t] 
    \centering
    \includegraphics[width=\linewidth]{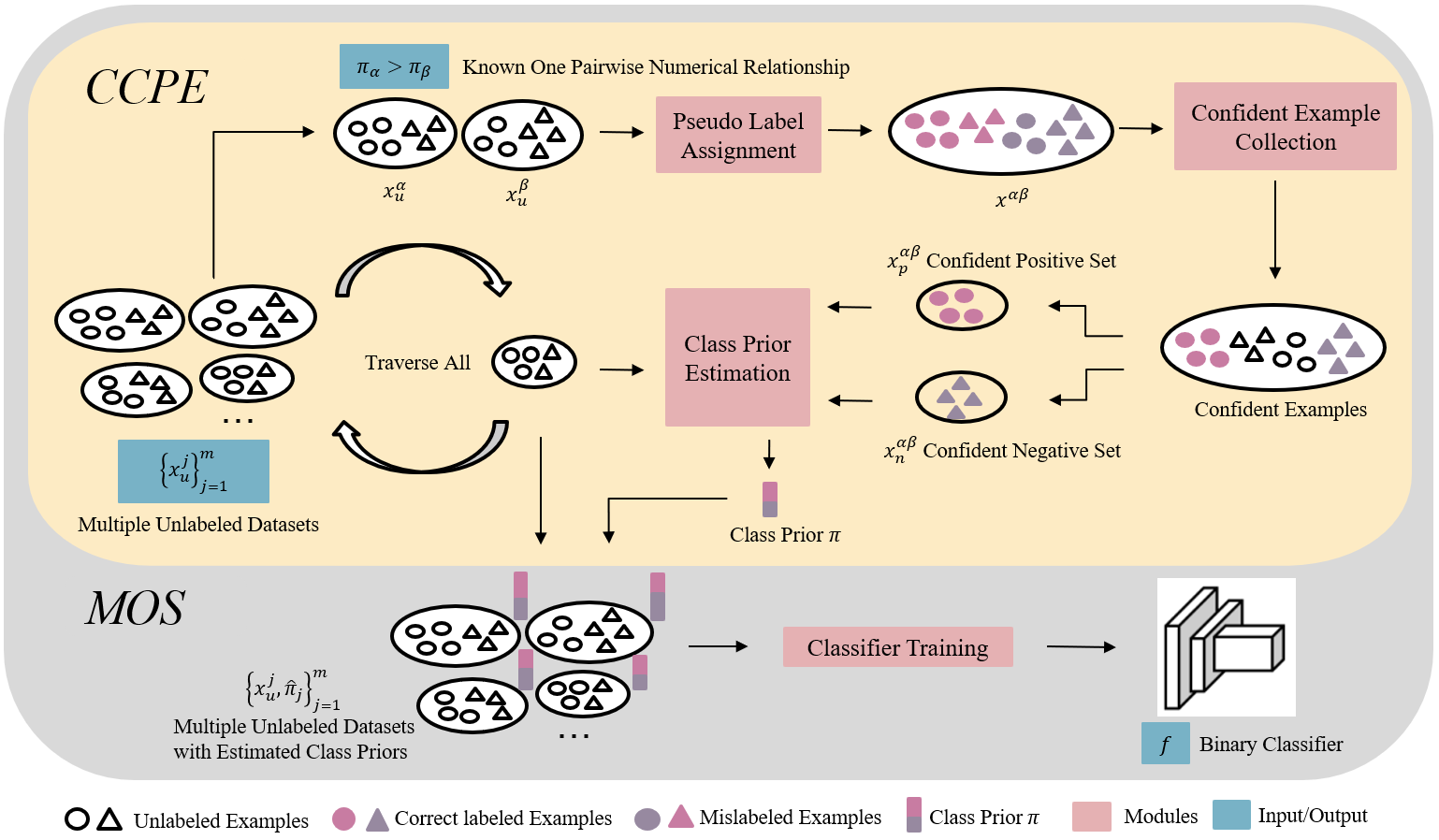}
    \begin{center}
  \caption{The illustration of tackling the MU-OPPO problem with our \underline{M}U-\underline{O}PPO \underline{s}olution (MOS) framework. The proposed MOS framework contains four modules: (i) pseudo label assignment; (ii) confident example collection; (iii) class prior estimation; (iv) classifier training with estimated class priors. The first three modules mainly constitute the \underline{c}onfident \underline{c}lass-\underline{p}rior \underline{e}stimator (CCPE) to estimate multiple class priors by utilizing only one pairwise numerical relationship of class priors. Then, the estimated class priors are employed to build statistically consistent algorithms for binary classification from unlabeled datasets in the fourth module. The CCPE is included in the MOS framework.}
    \label{illustration}
    \end{center}
    \vspace{-2em}
\end{figure*}

In this section, we design a solution framework to handle the MU-OPPO problem and name it the \underline{M}U-\underline{O}PPO \underline{s}olution (MOS) framework. The proposed MOS framework of the MU-OPPO problem contains four modules: (i) pseudo label assignment; (ii) confident example collection; (iii) class prior estimation; (iv) classifier training with estimated class priors. The first three modules estimate the class priors and are included in the CCPE. The entire four modules mainly constitute the MOS framework and tackle the MU-OPPO problem sequentially. The illustration of tackling the MU-OPPO problem is provided in Figure~\ref{illustration}. Specifically, we first assign pseudo labels to two unlabeled datasets, whose numerical relationship between class priors is known and collect confident examples used to approximate the latent distributions $\mathbbm{P}_\mathrm{p}$ and $\mathbbm{P}_\mathrm{n}$. Then, with the selected confident examples, the class priors of all unlabeled datasets are estimated. Finally, the estimated class priors are employed to learn a binary classifier from multiple unlabeled datasets. Note that our MOS framework consists of four modules that provide compatibility with some existing methods, e.g., adopting various existing methods for selecting confident examples into our confident example collection module. In summary, we target a novel MU-OPPO problem setting by designing a new solution framework here. Technically, we propose a new way of estimating multiple class priors with only one pairwise numerical relationship of class priors. Below we discuss the technical details of the modules involved in our MOS framework.

\subsection{Pseudo Label Assignment}
We first discuss how to assign pseudo labels to multiple unlabeled datasets. Specifically, for $\alpha,\beta\in\{1,\ldots,m\}$, we build a new dataset by combining the two unlabeled datasets, i.e., 
\begin{equation}
\mathcal{X}_{\mathrm{u}}^{\alpha} = \{\bm{x}_{1}^{\alpha},\ldots, \bm{x}_{\mathrm{n}_\alpha}^{\alpha}\} \overset{\rm{i.i.d.}}{\sim} \mathbbm{P}_{\mathrm{u}}^{{\alpha}}(\boldsymbol{x})\ \text{and} \  \mathcal{X}_{\mathrm{u}}^{\beta} = \{\bm{x}_{1}^{\beta},\ldots, \bm{x}_{\mathrm{n}_\beta}^{\beta}\} \overset{\rm{i.i.d.}}{\sim} \mathbbm{P}_{\mathrm{u}}^{{\beta}}(\bm{x}). 
\end{equation}
Afterwards, pseudo labels $\tilde{y}_i^\alpha$ and $\tilde{y}_i^\beta$ corresponding to $\bm{x}_i^\alpha$ and $\bm{x}_i^\beta$ are assigned with positive~($+1$) and negative~($-1$) forms. 

The numerical relationship of class priors $\pi_\alpha$ and $\pi_\beta$ play an important role in assigning pseudo labels. In detail, if $\pi_\alpha >\pi_\beta$, we can regard $\mathbbm{P}_\mathrm{u}^\alpha(\bm{x})$ as the corrupted positive density $\tilde{\mathbbm{P}}_\mathrm{p}^\alpha(\bm{x})$, and regard $\mathbbm{P}_\mathrm{u}^\beta(\bm{x})$ as the corrupted negative density $\tilde{\mathbbm{P}}_\mathrm{n}^\beta(\bm{x})$. Note that we assume that among the $m$ unlabeled datasets, at least two class priors are different and their numerical relationship is known. We use the pair that has different class priors. Therefore, we roughly assign positive labels to $\mathcal{X}_\mathrm{u}^\alpha$ and negative labels to $\mathcal{X}_\mathrm{u}^\beta$. After the assignment of pseudo labels, we obtain a pseudo labeled dataset $\mathcal{X}^{\alpha\beta}=\{(\bm{x}_i,\tilde{y}_i)\}_{i=1}^{n_\alpha+n_\beta}$, where $\tilde{y}_i$ denotes the (noisy) pseudo labels of the instance $\bm{x}_i$.

At first glance, the generation process of our pseudo-labeled dataset looks similar to class-conditional noise~\citep[CCN,][]{angluin1988learning} in learning with noisy labels~\cite{Liu_2016,natarajan2013learning}, where clean labels are assumed to flip into other classes with a certain probability. In fact, this data generation process is highly related to the mutually contaminated distributions~\citep[MCD,][]{scott2013classification,ICLR:Lu+etal:2019}, which is more general than the CCN model. Denote $\tilde{\mathbbm{P}}(\cdot)$ by the corrupted distribution and $\tilde{y}$ by the random variable of noisy labels. Then, CCN and MCD are defined by

\begin{align*}
    \begin{pmatrix}
    \tilde{\mathbbm{P}}(\tilde{y}=+1\mid \bm{x})\\
    \tilde{\mathbbm{P}}(\tilde{y}=-1\mid \bm{x})
    \end{pmatrix}
    = T_{\textrm{CCN}}
    \begin{pmatrix}
    \mathbbm{P}(y=+1\mid \bm{x})\\
    \mathbbm{P}(y=-1\mid \bm{x})
    \end{pmatrix}
    \quad\textrm{and}\quad
    \begin{pmatrix}
    \tilde{\mathbbm{P}}(\bm{x}\mid\tilde{y}=+1)\\
    \tilde{\mathbbm{P}}(\bm{x}\mid\tilde{y}=-1)
    \end{pmatrix}
    = T_{\textrm{MCD}}
    \begin{pmatrix}
    \mathbbm{P}_{\mathrm{p}}(\bm{x})\\
    \mathbbm{P}_{\mathrm{n}}(\bm{x})
    \end{pmatrix},
\end{align*}
where both of $T_{\textrm{CCN}}$ and $T_{\textrm{MCD}}$ are $2\times 2$ matrices but $T_{\textrm{CCN}}$ is column normalized and $T_{\textrm{MCD}}$ is row normalized. Moreover, the CCN model is a strict special case of the MCD model~\cite{menon2015learning}. Denote $\tilde{\mathbbm{P}}(\tilde{y})$ by the corrupted label distribution. For CCN, $\tilde{\mathbbm{P}}(\tilde{y})$ is fixed once $\tilde{\mathbbm{P}}(\tilde{y}|\bm{x})$ is specified. While, for MCD, $\tilde{\mathbbm{P}}(\tilde{y})$ is free after $\mathbbm{P}(\bm{x}|
\tilde{y})$ is specified. Furthermore, $\tilde{\mathbbm{P}}(\bm{x})=\mathbbm{P}(\bm{x})$ in CCN but $\tilde{\mathbbm{P}}(\bm{x})\neq\mathbbm{P}(\bm{x})$ in MCD. Due to this covariate shift~\cite{sugiyama2012machine}, CCN methods do not fit the MCD problem setting, while the MCD methods can fit the CCN problem setting conversely~\cite{wu2022learning}.

After generating the pseudo-labeled dataset, we warm up the training of a binary classifier, e.g., ResNet~\cite{he2016deep}, in the first few epochs to help identify confident examples later. The reason for training a warm-up classifier is that the deep learning models will first memorize training data with clean labels and then gradually adapt to those with noisy labels as training epochs become large~\cite{han2018co,xia2020robust,arpit2017closer,jiang2018mentornet,pleiss2020identifying}. Therefore, the classifier's predictions can be reliable after a few warm-up learning epochs. 

\subsection{Confident Example Collection} \label{Confident-Example Collection module}

In order to approximate the latent distributions, i.e., $\mathbbm{P}_\mathrm{p}$ and $\mathbbm{P}_\mathrm{u}$, we collect confident examples from $\mathcal{X}^{\alpha\beta}$ based on a previously trained warm-up classifier. For collecting confident examples, some existing methods in the literature of learning with label noise have been proven to work well. Here, we adopt three representative methods among them from different perspectives: (i) the loss distribution~\cite{arazo2019unsupervised}; (ii) the prediction probability of examples on given pseudo labels~\cite{northcutt2021confident}; (iii) the latent representation of data points ~\cite{kim2021fine}. In the following, confident examples are formally defined first, and then we discuss three methods for collecting confident examples through the trained warm-up classifier.

\begin{definition}[Confident Example] \label{confident examples}

$\mathcal{D}$ is the underlying joint distribution of a pair of random variables $(\bm{x},y) \in \mathcal{X} \times \mathcal{Y}$. For any example $(\bm{x}_i,\tilde{y}_i)$ sampled from the joint distribution $\mathcal{D}$, if the probability 
\begin{equation} \label{confident_example}
    \mathbbm{P}(y=\tilde{y}_i\mid\bm{x}=\bm{x}_i) > 0.5,
\end{equation}
the example $(\bm{x}_i,\tilde{y}_i)$ is defined as a confident example.  
\end{definition}

In the above definition, Eq.~(\ref{confident_example}) denotes that an example can be defined as a confident example when the probability that its assigned pseudo label is correct is greater than the threshold of $0.5$. In the following implementations, we collect confident examples by estimating the posterior probability $\mathbbm{P}(y=\tilde{y}_i\mid\bm{x}=\bm{x}_i)$. It is inevitable that estimation errors will occur. Thus, in practice, we appropriately increase the threshold to collect confident samples more strictly.   

\subsubsection{Collection by the Loss Distribution}
It has been empirically demonstrated~\cite{han2018co,song2019does} that deep networks tend to memorize clean examples faster than mislabeled examples. Thus, when considering the loss value of examples after the warm-up training, clean examples are more likely to have smaller losses than mislabeled examples~\cite{han2018co,chen2019understanding}. Following~\cite{arazo2019unsupervised}, we estimate the probability that an example is confident by fitting a two-component mixture model to the loss distribution for mixtures of clean and mislabeled examples. Formally, let $\bm{\theta}_w$ denote the parameters of our warm-up classifier. The logistic loss $\mathcal{L}(\bm{\theta}_{w})$ reflects how well the classifier fits our pseudo labeled dataset $\mathcal{X}^{\alpha\beta}$:

\begin{equation}
\mathcal{L}(\bm{\theta}_{w})=\{\mathcal{L}_i\}^{n_\alpha+n_\beta}_{i=1}=\{\ln(1 + \exp^{-\tilde{y}_i f(\bm{x}_i; \bm{\theta}_{w})}) \}^{n_\alpha+n_\beta}_{i=1},
\end{equation}
where $f(\bm{x}_i;\theta_{w})$ denotes the predicted probability by the warm-up classifier when the pseudo label is $\tilde{y}_i$.

Considering that the Gaussian Mixture Model (GMM) can better distinguish clean and mislabeled samples due to its flexibility in the sharpness of distribution~\cite{permuter2006study}, we fit a two-component GMMs with $\mathcal{L}(\bm{\theta}_{w})$ using the Expectation-Maximization algorithm~\cite{JRSS-B:Dempster+etal:1977}. For each example, the probability $w_i$ that the assigned pseudo label is correct is estimated by the posterior probability $\mathbbm{P}(g|\bm{x}_i)$, where $g$ is the Gaussian component with a smaller mean (minor loss). We collect confident examples by setting a threshold on $w_i$.

\subsubsection{Collection by the Prediction Probability}

The confident learning algorithm~\cite{northcutt2021confident} works by estimating the joint distribution of noisy and latent true labels. The estimation relies on the predicted probability of examples on given pseudo labels. The central idea of the confident learning algorithm is to introduce the \emph{confident joint} $C_{\tilde{y},y}$ to partition and count confident examples. 

The confident joint $C_{\tilde{y},y}$ first finds out the set of examples with the noisy label $r$ and true label $s$, which is denoted by $\mathcal{X}_{\tilde{y}=r,y=s}$. Afterward, the algorithm identifies $\mathcal{X}^*_{\tilde{y}=r,y=s}$ out of $\mathcal{X}_{\tilde{y}=r,y=s}$, where $\mathcal{X}^*_{\tilde{y}=r,y=s}$ is the set of examples noisily labeled $\tilde{y}=r$ with a large enough predicted probability $\hat{\mathbbm{P}}(\tilde{y}=s|\bm{x})$. Here, the predicted probability is output by our trained warm-up classifier. Moreover, by introducing a per-class threshold $t_s$, the confident joint is formulated as 

\begin{equation}
    C_{\tilde{y},y}[r][s] \coloneqq \mathcal{X}^*_{\tilde{y}=r,y=s},
\end{equation}
where 
\begin{equation}
\mathcal{X}^*_{\tilde{y}=r,y=s} \coloneqq \left\{ \bm{x} \in \mathcal{X}_{\tilde{y} = r} : \; \hat{\mathbbm{P}} (\tilde{y} = s |\bm{x}) \ge t_s, \,\, s = \argmax \hat{\mathbbm{P}} 
(\tilde{y} = l |\bm{x})  \right\}.
\end{equation}
The threshold $t_s$ is the expected (average) self-confidence in each class, i.e., 
\begin{equation} 
     t_s = \frac{1}{|\mathcal{X}_{\tilde{y}=s}|} \sum_{\bm{x} \in \mathcal{X}_{\tilde{y}=s}} \hat{\mathbbm{P}}(\tilde{y}=s|\bm{x}).
\end{equation}

It should be noted that in the original paper of the confident learning algorithm~\cite{northcutt2021confident}, the computation of predicted probabilities is \emph{out-of-sample} by using four-fold cross-validation. Otherwise, the predicted probabilities of examples would be \emph{over-confident}\footnote{
When training neural networks on train datasets, modern neural networks are over-confident in their predictions, i.e., the predicted probabilities of examples are excessively high~\cite{wang2021confident,guo2017calibration}.}. Fortunately, using the warm-up classifier in our framework has already addressed this issue and would be reliable enough for the downstream collection of confident examples, as experimentally demonstrated in Section~\ref{sec:6}.

After computing $C_{\tilde{y},y}$, we could easily collect confident examples $\{\bm{x}\in \hat{\mathcal{X}}_{\tilde{y}=r,y=s} \colon r = s\}$ from the diagonals of $C_{\tilde{y},y}$. Specifically, we can obtain the confident positive set $\mathcal{X}^{\alpha\beta}_\mathrm{p}$ and confident negative set $\mathcal{X}^{\alpha\beta}_\mathrm{n}$ as
\begin{equation}
    \mathcal{X}^{\alpha\beta}_\mathrm{p}=\{\bm{x}\in \hat{\mathcal{X}}_{\tilde{y}=+1,y=+1}\}\ \text{and} \ \mathcal{X}^{\alpha\beta}_\mathrm{n}=\{\bm{x}\in \hat{\mathcal{X}}_{\tilde{y}=-1,y=-1}\}.
\end{equation}

\subsubsection{Collection by the Latent Representation} 

In~\cite{kim2021fine}, the confident examples are identified by the alignment between the principal component distribution and representation of each instance by using an eigenvalue decomposition of the data Gram matrix. For this goal, given the low-dimensional representation $\bm{z}_i$ of the instance $\bm{x}_i$, which can be obtained with our previously trained warm-up classifier, the data Gram matrix is defined as 
\begin{equation}
    \bm{G}_{k} = \sum_{\tilde{y}_i:= k}\bm{z}_i\bm{z}_i^\top, k\in\{+1,-1\}.
\end{equation}
Then, the alignment $\bm{a}_i$ of $\bm{x}_i$ is evaluated via the square of the inner product, i.e., 
\begin{equation}\label{alignment}
\bm{a}_{i,k}:={\langle \bm{u}_k,\bm{z}_i\rangle}^2.
\end{equation}
Here, $\bm{u}_k$ is the first column of $\bm{U}_k$ from the eigendecomposition of $\bm{G}_{k}$, when the eigenvalues are in descending order. According to \cite{kim2021fine}, the collection process of confident examples depends on \textit{alignment clusterability}, which leads to the following definition. 

\begin{definition}[Alignment Clusterability~\cite{kim2021fine}]\label{AlignmentClusterability} 
For all features pseudo-labeled as class $k$ in the dataset $\mathcal{X}^{\alpha\beta}$, let fit a two-component GMM on their alignment $\bm{a}_{i,k}$ to divide them into two sets. The set that has a larger mean value is treated as a confident set. Then, we state that the dataset $\mathcal{X}^{\alpha\beta}$ satisfies alignment clusterability if the representation $\bm{z}$ labeled as the same true class belongs to the confident set.
\end{definition}
As a whole, $\bm{u}_k$ is the principal component with the eigendecomposition. The dataset is well clustered with the alignment of representations toward the principal component. Hence, for the dataset $\mathcal{X}^{\alpha\beta}$, we can obtain the confident positive set $\mathcal{X}^{\alpha\beta}_\mathrm{p}$ and confident negative set $\mathcal{X}^{\alpha\beta}_\mathrm{n}$ by fitting the GMM on their alignment.

Till now, we have discussed how to collect confident examples. Below, we show how to estimate the class priors accurately with these confident examples.

\subsection{Class Prior Estimation} \label{Class-Prior estimate module}

According to Section~\ref{sec:2.2}, the direct application of MPE-based class prior estimators is infeasible due to the latent $\mathbbm{P}_\mathrm{p}$ or $\mathbbm{P}_\mathrm{n}$ distribution. But, the above confident example collection module can approximate the latent distribution by selecting confident samples. Thus, we can estimate class priors in our setting by implementing some existing methods on the MPE problem. Note that without any assumption, the class priors are not identifiable in the MPE problem~\cite{blanchard2010semi,scott2015rate,yao2022rethinking}. Thus, to ensure identifiability, the irreducible assumption~\cite{blanchard2010semi} has been proposed, and we briefly review this assumption first in this subsection. After introducing the irreducible assumption, we provide how to estimate class priors in our setting using a standard MPE estimator when the confident examples are collected. Later, we also implement two recent MPE estimators, i.e., Regrouping MPE (ReMPE), which weakens the irreducible assumption~\cite{yao2022rethinking}, and the Best
Bin Estimation (BBE) estimator~\cite{garg2021mixture}.

To introduce the irreducible assumption, let $\mathbbm{P}_\mathrm{p}$ and $\mathbbm{P}_\mathrm{u}$ be probability distributions on the measurable space $(\mathcal{X},\Omega)$, where $\Omega$ is the $\sigma$-algebra. Let $\kappa^{*}$, which is the maximum proportion of $\mathbbm{P}_\mathrm{p}$ in $\mathbbm{P}_\mathrm{u}$, be identical to $\pi$. The irreducible assumption was proposed by~\cite{blanchard2010semi} as follows:

\begin{assumption}[Irreducible]\label{ass:irr}
The distribution $\mathbbm{P}_\mathrm{n}$ is irreducible with respect to $\mathbbm{P}_\mathrm{p}$, if $\mathbbm{P}_\mathrm{n}$ is not a mixture containing $\mathbbm{P}_\mathrm{p}$. That is, it is not possible to admit a decomposition $\mathbbm{P}_\mathrm{n}=(1-\gamma)Q+\gamma\mathbbm{P}_\mathrm{p}$, where $Q$ is a probability distribution on the measurable space $(\mathcal{X},\Omega)$, and $0 < \gamma \leq 1$.
\end{assumption}

Assumption~\ref{ass:irr} implies the following fact:
\begin{equation}
\inf_{S \in \Omega, \mathbbm{P}_\mathrm{p}(S)>0}\frac{\mathbbm{P}_{\mathrm{n}}(S)}{\mathbbm{P}_\mathrm{p}(S)} = 0.
\end{equation}
This means that with the selection of different sets $S$, the probability $\mathbbm{P}_\mathrm{n}(S)$ can be arbitrarily close to 0, and $\mathbbm{P}_\mathrm{p}(S)>0$~\cite{scott2015rate,yao2022rethinking}. Intuitively, the irreducible assumption assumes that the support of the positive class-conditional
distribution $\mathbbm{P}_\mathrm{p}$ is not contained in the support of the negative class-conditional distribution $\mathbbm{P}_\mathrm{n}$.

\subsubsection{The Standard MPE Estimator}

Here, we introduce how to estimate class priors through the standard MPE estimator. Based on Assumption~\ref{ass:irr}, this standard MPE estimator estimates class priors with theoretical guarantees~\cite{scott2015rate}. Note that we can approximate both the latent $\mathbbm{P}_\mathrm{p}$ and the latent $\mathbbm{P}_\mathrm{n}$ distributions here by previously collected confident examples. Thus, by approximating these two latent distributions, it is possible to estimate class priors twice through the standard MPE estimator. In the following, we first introduce the standard MPE estimator and then describe how to estimate class priors twice with a suitable assumption to obtain more accurate results.

Based on Assumption~\ref{ass:irr}, a standard estimator can be designed with theoretical guarantees to estimate underlying class priors. Now we can access the distribution $\mathbbm{P}^j_\mathrm{u}$ and approximate the distribution $\mathbbm{P}_\mathrm{p}$, and suppose the set $C$ contains all possible latent distributions. The class prior $\pi_j$ can be estimated as 

\begin{equation}
\label{estimator1}
    \hat{\pi}_j^1 = \kappa^{*}(\mathbbm{P}_\mathrm{u}^j|\mathbbm{P}_\mathrm{p}) := \sup\{\omega|\mathbbm{P}_{\mathrm{u}}^j = (1-\omega)K + \omega\mathbbm{P}_{\mathrm{p}}, K \in C\} 
     =\inf_{S \in \Omega, \mathbbm{P}_\mathrm{p}(S)>0}\frac{\mathbbm{P}_{\mathrm{u}}^j(S)}{\mathbbm{P}_{\mathrm{p}}(S)}.
\end{equation}

According to~\cite{scott2015rate}, the above equation converges to the true class priors at an acceptable rate. However, approximating the distribution $\mathbbm{P}_\mathrm{p}$ requires the collection of confident samples, which involves errors in selecting confident examples. Thus, the mistakes of selecting confident examples cause the estimation errors of class priors, and we clearly show it in Section~\ref{sec:5.1}. To reduce the estimation errors of the above equation, we additionally estimate class priors using the approximation of $\mathbbm{P}_\mathrm{n}$ and average the estimation results. This is because we can additionally approximate the distribution $\mathbbm{P}_\mathrm{n}$ by our confident example collection module.

To additionally estimate class priors through the standard MPE estimator, the distribution $\mathbbm{P}_\mathrm{n}$ also needs to be irreducible with respect to $\mathbbm{P}_\mathrm{p}$. Thus, we employ the mutually irreducible assumption~\cite{scott2015rate} here, and then present the additional estimation formula.

\begin{assumption}[Mutually Irreducible]\label{ass:m_irr}
The distributions $\mathbbm{P}_\mathrm{p}$ and $\mathbbm{P}_\mathrm{n}$ are said to be mutually irreducible if $\mathbbm{P}_\mathrm{n}$ is irreducible with respect to $\mathbbm{P}_\mathrm{p}$, and vice versa.
\end{assumption}
Assumption~\ref{ass:m_irr} means that the support of $\mathbbm{P}_\mathrm{n}$ is also hardly contained in the support of $\mathbbm{P}_\mathrm{p}$: $\texttt{Supp}(\mathbbm{P}_\mathrm{p})\not\subset \texttt{Supp}(\mathbbm{P}_\mathrm{n})$ and $\texttt{Supp}(\mathbbm{P}_\mathrm{n})\not\subset \texttt{Supp}(\mathbbm{P}_\mathrm{p})$. Assumption~\ref{ass:m_irr} is reasonable in many cases, which essentially says that the existing patterns belonging to positive examples could not possibly be confused with patterns from negative examples. We thus have the following additional estimation formula:

\begin{equation}
\label{estimator2}
    \hat{\pi}_j^2 = 1 - \kappa^{*}(\mathbbm{P}_\mathrm{u}^j|\mathbbm{P}_\mathrm{n}):= 1- \inf_{S\in \Omega, \mathbbm{P}_\mathrm{p}(S)>0}\frac{\mathbbm{P}_{\mathrm{u}}^j(S)}{\mathbbm{P}_{\mathrm{n}}(S)}.
\end{equation}
By combining both Eq.~(\ref{estimator1}) and Eq.~(\ref{estimator2}), at last, the class prior $\pi_j$ can be estimated as $\hat{\pi}_j=(\hat{\pi}_j^1+\hat{\pi}_j^2)/2$. 

Note that, for estimating the class priors, most other previous methods, e.g., kernel mean embedding-based estimators KM1 and KM2~\cite{ramaswamy2016mixture}, a non-parametric class prior estimator AlphaMax (AM)~\cite{jain2016nonparametric}, Elkan-Noto (EN)~\cite{elkan2008learning}, DEDPUL~(DPL) \cite{ivanov2019dedpul}, and Rankprunning~(RP)~\cite{northcutt2017learning} are based on the standard MPE estimator and implicitly or explicitly rely on the irreducible assumption. However, this assumption is strong and hard to guarantee~\cite{yao2022rethinking} since the multiple unlabeled sets may be collected from diverse sources. Therefore, we apply the ReMPE estimator~\cite{yao2022rethinking} that improves the estimations of class priors without the irreducible assumption. 

\subsubsection{The ReMPE Estimator}

The main idea of the ReMPE estimator is that, instead of estimating the maximum proportion of $\mathbbm{P}_\mathrm{p}$ in $\mathbbm{P}_\mathrm{u}$, and we raise a new MPE problem by creating a new auxiliary distribution $\mathbbm{P}_{\mathrm{p}'}$. The new auxiliary distribution makes the irreducible assumption hold~\cite{yao2022rethinking}. Then we use the standard MPE estimator to obtain the class priors $\hat{\pi}_j$. We use the following regrouping process to construct the auxiliary distribution $\mathbbm{P}_{\mathrm{p}'}$.

The process of regrouping is to change the $\mathbbm{P}_\mathrm{n}$ and $\mathbbm{P}_\mathrm{p}$ into new $\mathbbm{P}_{\mathrm{n}'}$ and $\mathbbm{P}_{\mathrm{p}'}$ by transporting a small set of examples $A$ from $\mathbbm{P}_{\mathrm{n}}$ to $\mathbbm{P}_\mathrm{p}$. Note that, in the original paper~\cite{yao2022rethinking}, the set $A$ is selected from $\mathbbm{P}_{\mathrm{u}}$ due to the unavailability of $\mathbbm{P}_{\mathrm{n}}$ in the positive-unlabeled learning setting~\cite{denis2005learning}. Nevertheless, in our setting, we could directly collect $A$ from the distribution $\mathbbm{P}_{\mathrm{n}}$ that was approximated by selected confident negative sets $\mathcal{X}^{\alpha\beta}_\mathrm{n}$. 

After the regrouping process, the selected set $A$ should contain the examples that look the most similar to $\mathbbm{P}_\mathrm{p}$ and dissimilar to $\mathbbm{P}_{\mathrm{n}}$, which could guarantee better class prior estimation when irreducible assumption does not hold~\cite{yao2022rethinking}. Therefore, we train the binary classifier with the confident positive set $\mathcal{X}^{\alpha\beta}_\mathrm{p}$ and confident negative set $\mathcal{X}^{\alpha\beta}_\mathrm{n}$. Then, we obtain $\mathcal{X}^{\alpha\beta}_{\mathrm{p}'}$ by copying the percentage $p$ of $\mathcal{X}^{\alpha\beta}_\mathrm{n}$ examples with the smallest negative class-posterior probability from $\mathcal{X}^{\alpha\beta}_\mathrm{n}$ to $\mathcal{X}^{\alpha\beta}_\mathrm{p}$. In the end, we estimate the class priors by employing an algorithm based on Eq.~(\ref{estimator1}) with inputs $\mathcal{X}^{\alpha\beta}_{\mathrm{p}'}$ and $\mathcal{X}_{\mathrm{u}}^{j}$. The above process could be easily extended to obtain $\mathcal{X}^{\alpha\beta}_{\mathrm{n}'}$, which can be incorporated with $\mathcal{X}_{\mathrm{u}}^{j}$ on Eq.~(\ref{estimator2}) to finally estimate the class prior $\hat{\pi}_j$.

\subsubsection{The BBE Estimator}

Recently, \cite{garg2021mixture} proposed the Best Bin Estimation (BBE) for the MPE problem. This method produces a consistent class-prior estimation under the \emph{pure positive bin} assumption. The assumption means that the top bin (nearly) purely contains positive examples when packing unlabeled data into bins relying on their predicted probabilities (of being positive) from the trained Positive-versus-Unlabeled (PvU) classifier.  This assumption is a variation of the \emph{irreducible} assumption and sufficient for BBE to obtain a (nearly) consistent class-prior estimation~\cite{garg2021mixture}. 
Next, we summarize the main process of the BBE estimator in our setting, referring to~\cite{garg2021mixture}.

At the beginning, we train a Positive-versus-Unlabeled (PvU) classifer $f_{\mathrm{p},\mathrm{u}}$ on some portions of $\mathcal{X}^{\alpha\beta}_{\mathrm{p}}$ and $\mathcal{X}^{j}_{\mathrm{u}}$ and push other positive and unlabeled
examples through the classifier $f_{\mathrm{p},u}$ to obtain one-dimensional outputs $Z_{\mathrm{p}} = f_{\mathrm{p},u}(\mathcal{X}^{\alpha\beta}_{\mathrm{p}})$ and $Z_{\mathrm{u}}^{j} = f_{\mathrm{p},u}(\mathcal{X}^{j}_{\mathrm{u}})$. We now define a function $q(z)=\int_{H_z} p(x)dx$, where $H_{z} = \{x \in \mathcal{X}: f(x) 	\geq z \}$ for all $z \in [0,1]$. Intuitively, $q(z)$ captures the cumulative density of points in a top bin, i.e., the proportion of the input domain is assigned a value larger than $z$ by the classifier $f$ in the transformed space. With the $Z_{\mathrm{p}}$ and $Z_{\mathrm{u}}^{j}$, the estimations of $q_{\mathrm{p}}(z)$ and $q_{\mathrm{u}}^{j}(z)$ are

\begin{equation}
    \hat{q}_\mathrm{p} (z) = \frac{\sum_{z_i \in  Z_\mathrm{p}} \mathbbm{1}[z_i \ge z]}{n_\mathrm{p}}\quad\text{and}\quad\hat{q}_\mathrm{u}^j(z)= \frac{\sum_{z_i \in  Z_\mathrm{u}^j} \mathbbm{1}[z_i \ge z]}{n_\mathrm{u}^j}\quad \text{ for all } z \in [0,1].
\end{equation}
Then, we obtain $\hat{c}$ that minimizes the upper confidence bound as follows:

\begin{equation}
\label{BBE}
\hat{c}= \argmin_{c \in [0,1]} \left( \frac{\hat{q}_{\mathrm{u}}^j(c)}{\hat{q}_\mathrm{p}(c)}  + \frac{1+\gamma}{\hat{q}_\mathrm{p}(c)}\left( \sqrt{\frac{\log(4/\delta)}{2 n_\mathrm{u}^j}} + \sqrt{\frac{\log(4/\delta)}{2n_\mathrm{p}}}\right) \right)\,,
\end{equation}
at a pre-specified level $\gamma$ and a fixed parameter $\delta \in (0,1)$. The above Eq.~(\ref{BBE}) has been theoretically proven and empirically verified in \cite{garg2021mixture}. Finally, we obtain the class-prior estimation $\hat{\pi}_{j}^{1} = \hat{q}_\mathrm{u}^j(\hat{c})/\hat{q}_\mathrm{p}(\hat{c})$. Furthermore, $\hat{\pi}_{j}^{2}$ also could be similarly estimated by replacing $\mathcal{X}^{\alpha\beta}_{\mathrm{p}}$ as $\mathcal{X}^{\alpha\beta}_{\mathrm{n}}$ in the above process. Accordingly, the class prior $\pi_j$ can be estimated as $\hat{\pi}_j=(\hat{\pi}_j^1+\hat{\pi}_j^2)/2$ by the BBE estimator. 

Till now, the first three modules make up the proposed CCPE that can estimate all class priors by utilizing only one pairwise numerical relationship of class priors. We summarize the algorithm flow of CCPE in Algorithm~\ref{Confident-MPE}. Note that since the approximation of $\mathbbm{P}_\mathrm{p}$ and $\mathbbm{P}_\mathrm{n}$ only uses the selected confident examples from $\mathcal{X}^{\alpha\beta}$, the samples from other unlabeled datasets are wasted. Therefore, we construct the \underline{e}nhanced \underline{CCPE}~(ECCPE) that can substitute the CCPE to lead to a more accurate class-prior estimation. Specifically, in ECCPE, we run CCPE first and obtain the initialization of the class priors $\{\hat{\pi}_j\}_{j=1}^m$, which could identify the numerical relationships of all pairs of class priors. The size of the numerical relationships is $\tbinom{m}{2}$. After pairing the unlabeled datasets and selecting confident examples from all pairs of unlabeled datasets, class priors can be estimated multiple times according to different unlabeled dataset pairs. Thus, after averaging all estimated results of class priors, the final estimated results will be more accurate. We summarize the algorithm flow of ECCPE in Algorithm \ref{Unbiased-Confident-MPE}. 

\subsection{Classifier Training with Estimated Class Priors}\label{sec:3.4}

After obtaining the estimated class priors $\{\hat{\pi}_j^{\star}\}_{j=1}^m$ by one pairwise numerical relationship of class priors, for binary classifier training, an empirical risk minimization method can be constructed~\cite{lu2021binary}. In fact, other methods, e.g., \cite{scott2020learning}, can be applied with estimated class priors. We employ MCM~\cite{scott2020learning} and U$^m$-SSC~\cite{lu2021binary} here.

\subsubsection{MCM}

There have already been some risk-consistent methods for learning a binary classifier from only two unlabeled datasets with given class priors, e.g., BER~\cite{menon2015learning}, UU~\cite{ICLR:Lu+etal:2019}, and UU-c~\cite{lu2020mitigating}. Based on the risk-consistent methods, \cite{scott2020learning} proposed a mutual contamination (MCM) framework, which can learn a binary classifier from multiple unlabeled datasets in two steps: 

Firstly, pairing all unlabeled datasets so that they are sufficiently different in each pair. Specifically, MCM assumes the number of unlabeled datasets $m = 2k$ and conducts a pre-processing step that finds $k$ pairs of the unlabeled datasets by the index $t \in \{1, \ldots, k\}$. Let $(\mathcal{X}_{\rm{u}}^{t,+},\hat{\pi}_{t}^{\star,+})$ and $(\mathcal{X}_{\rm{u}}^{t,-},\hat{\pi}_{t}^{\star,-})$ constitute the $t$-th pair of bags such that the numerical relationship $\hat{\pi}_{t}^{\star,-} < \hat{\pi}_{t}^{\star,+}$ (MCM assumes that $\hat{\pi}_{t}^{\star,-} \neq \hat{\pi}_{t}^{\star,+}$ in each pair). The pairing of all unlabeled datasets maximizes 
\begin{equation}
\sum\nolimits^{k}_{t=1} \bar{n}_t (\hat{\pi}_{t}^{\star,+} - \hat{\pi}_{t}^{\star,-})^2,
\end{equation}
where $\bar{n}_t = 2n_t^{+}n_t^{-}/(n_t^{+} + n_t^{-})$ and $n_t^{+}$ and $n_t^{-}$ are the sizes of unlabeled dataset $\mathcal{X}_{\rm{u}}^{t,+}$ and $\mathcal{X}_{\rm{u}}^{t,-}$ respectively. For intuitive understanding, this pairing way gives preference to pairs of datasets where one dataset contains mostly positive samples (large $\hat{\pi}_{t}^{\star,+}$) and the other contains mostly negative samples (small $\hat{\pi}_{t}^{\star,-}$).

According to~\cite{scott2020learning}, the above pairing way is known as the ``maximum weighted matching'' problem with an exact solution~\cite{edmonds1965maximum}. Several approximation algorithms also exist for this problem. Note that when the sample sizes of $m$ unlabeled datasets are the same, the solution to this pairing problem is straightforward. That is, we can match the unlabeled dataset with the largest class prior $\hat{\pi}_{t}^{\star,+}$ to the unlabeled dataset with the smallest one and match the unlabeled dataset with the second largest class prior $\hat{\pi}_{t}^{\star,+}$ to the unlabeled dataset with the second smallest, and so on.

Secondly, after pairing all the unlabeled datasets, the unbiased risk estimators of each pair are linearly combined by the weights
\begin{equation}
w_t = \bar{n}_t (\hat{\pi}_{t}^{\star,+} - \hat{\pi}_{t}^{\star,-})^2.
\end{equation}
The resulting weighted learning objective is given by
\begin{equation}
\widehat{R}_\text{MCM}(f)=\sum\nolimits_{t=1}^{k}\omega_t\widehat{R}_{\rm{U^2}\text{-}\rm{c}}(f),
\end{equation}
where $\widehat{R}_{\rm{U^2}\text{-}\rm{c}}(f)$ is our selected non-negative risk estimator~\cite{lu2020mitigating} since it avoids overfitting during training on two unlabeled datasets and shows better empirical performance in the MCM framework~\cite{lu2021binary}.

\subsubsection{U$^m$-SSC}

\cite{lu2021binary} considered a surrogate set classification that bridges the original and surrogate class-posterior probabilities with a
linear-fractional transformation. Let the index of $\mathbbm{P}_\mathrm{u}^j$ be a surrogate label $\bar{y}\in\{1,\ldots,m\}$, $\bar{\mathcal{D}}$ be the joint distribution of $\bm{x}\in\mathcal{X}$ and $\bar{y}\in\bar{\mathcal{Y}}=\{1,\ldots,m\}$, $\eta(\bm{x})=\mathbbm{P}(y=+1|\bm{x})$ is the class-posterior probability for the true class +1 in the original binary classification, $\Bar{\eta}_j(\bm{x}) = \mathbbm{P}(\Bar{y} = j \mid \bm{x})$ is the class-posterior probability for the class $j$ in the surrogate set classification problem, and $\pi_{\mathcal{D}}$ is the test class prior. The goal of surrogate set classification is to train a classifier $g:\mathcal{X}\rightarrow \mathbb{R}^m$ that minimizes the following risk:
    \begin{align}
    \label{surrorisk}
        {R_{\rm{surr}}(g)=\mathbb{E}_{(\boldsymbol{x},\bar{y})\sim\bar{\mathcal{D}}}[\ell(g(\boldsymbol{x}),\bar{y})]},
    \end{align}
where $g(\bm{x})$ estimates the surrogate class-posterior probability $\bar{\eta}_j(\bm{x}) = \mathbbm{P}(\Bar{y} = j |\bm{x})$. Then, we bridge $\eta(\boldsymbol{x})$ and $\Bar{\eta}_j(\boldsymbol{x})$ by adding an estimated linear-fractional transition $\hat{T}_j$ with the final estimated class priors $\{\hat{\pi}_j^{\star}\}_{j=1}^m$, i.e., 
\begin{equation}
\Bar{\eta}_j(\boldsymbol{x})=\hat{T}_j(\eta(\boldsymbol{x})),\quad \forall j = 1,\ldots,m,
\end{equation}
where
\begin{equation} \nonumber
\hat{T}_j(\eta(\boldsymbol{x}))=\displaystyle\frac{\hat{a}_j\cdot\eta(\boldsymbol{x})+\hat{b}_j}{\hat{c}\cdot\eta(\boldsymbol{x})+\hat{d}},
\end{equation}
with $\hat{a}_j = {\rho}_j(\hat{\pi}_j^{\star}-\pi_{\mathcal{D}})$, $\hat{b}_j = {\rho}_j\pi_{\mathcal{D}}(1-\hat{\pi}_j^{\star})$, $\hat{c} = \sum_{j=1}^m{\rho}_j(\hat{\pi}_j^{\star}-\pi_{\mathcal{D}})$, and $\hat{d} = \sum_{j=1}^m{\rho}_j\pi_{\mathcal{D}}(1-\hat{\pi}_j^{\star})$. Here, $\rho_j$ is given by $\rho_j={n_j}/{\sum_{j=1}^{m}{n_j}}$. Afterwards, let $f(\bm{x})$ be the model outputs that estimate $\eta(\bm{x})$. We make use of the estimated transition function $\hat{T}_j(\cdot)$ and model $g_j(\bm{x})=\hat{T}_j\left(f(\boldsymbol{x})\right)$, where $g_j(\boldsymbol{x})$ is the $j$-th element of $g(\boldsymbol{x})$. Based on the above terms, the following modified loss function is presented as ${\ell}(g(\boldsymbol{x}), \bar{y})=\ell({{\hat{T}}}(f(\boldsymbol{x})), \bar{y})$, where ${\hat{T}}(f(\boldsymbol{x}))=[\hat{T}_1(f(\boldsymbol{x})),\ldots,\hat{T}_m(f(\boldsymbol{x}))]^\top$. 
Next, the corresponding risk for the surrogate task can be written as 
\begin{align}
R_{\rm{surr}}(f)=\mathbb{E}_{(\boldsymbol{x},\Bar{y})\sim\Bar{\mathcal{D}}}[\ell({\hat{T}}(f(\boldsymbol{x})), \bar{y})]=\mathbb{E}_{(\boldsymbol{x},\Bar{y})\sim\Bar{\mathcal{D}}}[{\ell}(g(\boldsymbol{x}), \bar{y})]= R_{\rm{surr}}(g).
\label{R_surr}
\end{align}
The corresponding empirical risk of Eq.~(\ref{R_surr}) is given by 
\begin{equation}\label{surrof}
        \widehat{R}_{\rm{surr}}(f)
        =\frac{1}{n'}\sum\nolimits_{i=1}^{n'}{\ell}({\hat{T}}(f(\boldsymbol{x}_i)),\bar{y}_i),
\end{equation}
where $n'$ denotes the number of all examples of multiple unlabeled datasets, i.e., $n'=\sum_{j=1}^m n_j$. \cite{lu2021binary} showed that the classiﬁer learned by solving the surrogate set classification task from multiple unlabeled datasets converges to the optimal classifier learned from fully supervised data under mild conditions. We empirically show that, in our setting, the nice theoretical properties are preserved. 

\begin{algorithm}[t]
    \caption{Algorithm flow of CCPE.}
    \label{Confident-MPE}
    \hspace*{0.02in} {\bf Input:} Unlabeled datasets 
    $\{{\mathcal{X}_{\rm{u}}^j}\}_{j=1}^{m}$ , which are known $\pi_{\alpha} > \pi_{\beta}$ for two unlabeled datasets $\mathcal{X}_\mathrm{u}^{\alpha}$ and $\mathcal{X}_\mathrm{u}^{\beta}$.
    \begin{algorithmic}[1]
        \STATE The pseudo labels $\tilde{y}_i^\alpha$ and $\tilde{y}_i^\beta$ corresponding to $\bm{x}_i^\alpha$ and $\bm{x}_i^\beta$ are assigned with +1 and -1.  \\
        \STATE Combining above items to obtain a pseudo labeled dataset $\mathcal{X}^{\alpha\beta}=\{(\bm{x}_i,\tilde{y}_i)\}_{i=1}^{n_\alpha+n_\beta}$.\\
        \STATE Collecting confident positive set $\mathcal{X}^{\alpha\beta}_\mathrm{p}$ and confident negative set $\mathcal{X}^{\alpha\beta}_\mathrm{n}$ out of $\mathcal{X}^{\alpha\beta}$.\\
        \FOR{$j = 1, 2,\ldots, m$}
            \STATE Estimating the class-prior $\hat{\pi}_j^{1}$ employing an algorithm based on Eq. (\ref{estimator1}) with inputs ${\mathcal{X}_{\rm{u}}^j}$ and $\mathcal{X}^{\alpha\beta}_\mathrm{p}$.
            \STATE Estimating the class-prior $\hat{\pi}_j^{2}$ employing an algorithm based on Eq. (\ref{estimator2}) with inputs ${\mathcal{X}_{\rm{u}}^j}$ and $\mathcal{X}^{\alpha\beta}_\mathrm{n}$.
            \STATE estimateing the class-prior $\hat{\pi}_j=(\hat{\pi}_j^1+\hat{\pi}_j^2)/2$.
        \ENDFOR    
    \end{algorithmic}
    \hspace*{0.02in} {\bf Output:} estimated class priors $\{\hat{\pi}_j\}_{j=1}^m$.
\end{algorithm}

\begin{algorithm}[t]
    \caption{ Algorithm flow of ECCPE}
    \label{Unbiased-Confident-MPE}
    \hspace*{0.02in} {\bf Input:} Unlabeled datasets $\{{\mathcal{X}_{\rm{u}}^j}\}_{j=1}^{m}$, which are known $\pi_{\alpha}>\pi_{\beta}$ for $\mathcal{X}_\mathrm{u}^{\alpha}$, $\mathcal{X}_\mathrm{u}^{\beta}$; Empty queue $Q$ and $T$; Pair-selection number $\gamma$.
    \begin{algorithmic}[1]
    \STATE Obtaining the initialization of the class priors $\{\hat{\pi}_j\}_{j=1}^m$ by running CCPE with $\{{\mathcal{X}_{\rm{u}}^j}\}_{j=1}^{m}$.\\
    \STATE Combining $\tbinom{m}{2}$ pairs of $\mathcal{X}_{\rm{u}}^{j}$ and identifying the numerical relationships of all pairs of class priors by $\{\hat{\pi}_j\}_{j=1}^m$.\\
    \STATE Re-indexing the unlabeled datasets by $t \in \{1, ..., \tbinom{m}{2}\}$ and let $(\mathcal{X}_{\rm{u}}^{t,+},\pi_{t}^{+})$ and $(\mathcal{X}_{\rm{u}}^{t,-},\pi_{t}^{-})$ constitute the $t$-th pair of bags, such that the numerical relationship: $\pi_{t}^{-} \leq \pi_{t}^{+}$, which has been identified above.\\
    \FOR{$t = 1, 2,\ldots, \tbinom{m}{2}$}
        \STATE Adding the value $\pi_{t}^{+} - \pi_{t}^{-}$ to queue $Q$ and $t$ to queue $T$.
    \ENDFOR
    \STATE Obtaining an indices list $q$ that would descending sort queue $Q$.
    \FOR{$k =1,2,\ldots,\gamma$}
    
        \STATE  Setting $t'$ equal to the $q[k]$-th element of $T$.
        \STATE  The pseudo labels $\tilde{y}_i^{+}$ and $\tilde{y}_i^{-}$ corresponding to $\mathcal{X}_{\rm{u}}^{t=t',+}$ and $\mathcal{X}_{\rm{u}}^{t=t',-}$ are assigned with +1 and -1. 
        \STATE Combining above items to obtain a pseudo labeled dataset $\mathcal{X}^{+-}=\{(\bm{x}_i,\tilde{y}_i)\}_{i=1}^{n_{+}+n_{-}}$.\\
        \STATE Collecting confident positive set $\mathcal{X}^{+-}_\mathrm{p}$ and confident negative set $\mathcal{X}^{+-}_\mathrm{n}$ out of $\mathcal{X}^{+-}$.\\
        \FOR{$j = 1, 2,\ldots, m$}
        \STATE Estimating the class-prior $\hat{\pi}_j^{1,t'}$ employing an algorithm based on Eq. (\ref{estimator1}) with inputs ${\mathcal{X}_{\rm{u}}^j}$ and $\mathcal{X}^{+-}_\mathrm{p}$.
        \STATE Estimating the class-prior $\hat{\pi}_j^{2,t'}$ employing an algorithm based on Eq. (\ref{estimator2}) with inputs ${\mathcal{X}_{\rm{u}}^j}$ and $\mathcal{X}^{+-}_\mathrm{n}$.
        \STATE estimateing the class-prior $\hat{\pi}_j^{t'}=(\hat{\pi}_j^{1,t'}+\hat{\pi}_j^{2,t'})/2$.
    \ENDFOR  
    \ENDFOR
    \STATE Obtaining final estimated class priors  $\{\hat{\pi}_j^{\star}\}_{j=1}^m = \{\frac{\sum^{t'} \hat{\pi}_j^{t'}}{\gamma}\}_{j=1}^m$.

    \end{algorithmic}
    \hspace*{0.02in} {\bf Output:} Final estimated class priors $\{\hat{\pi}_j^{\star}\}_{j=1}^m$.
\end{algorithm}

\section{Comparison with Related Work} \label{sec:4}

\begin{figure*}[!t] 
    \centering
    \includegraphics[width=\linewidth]{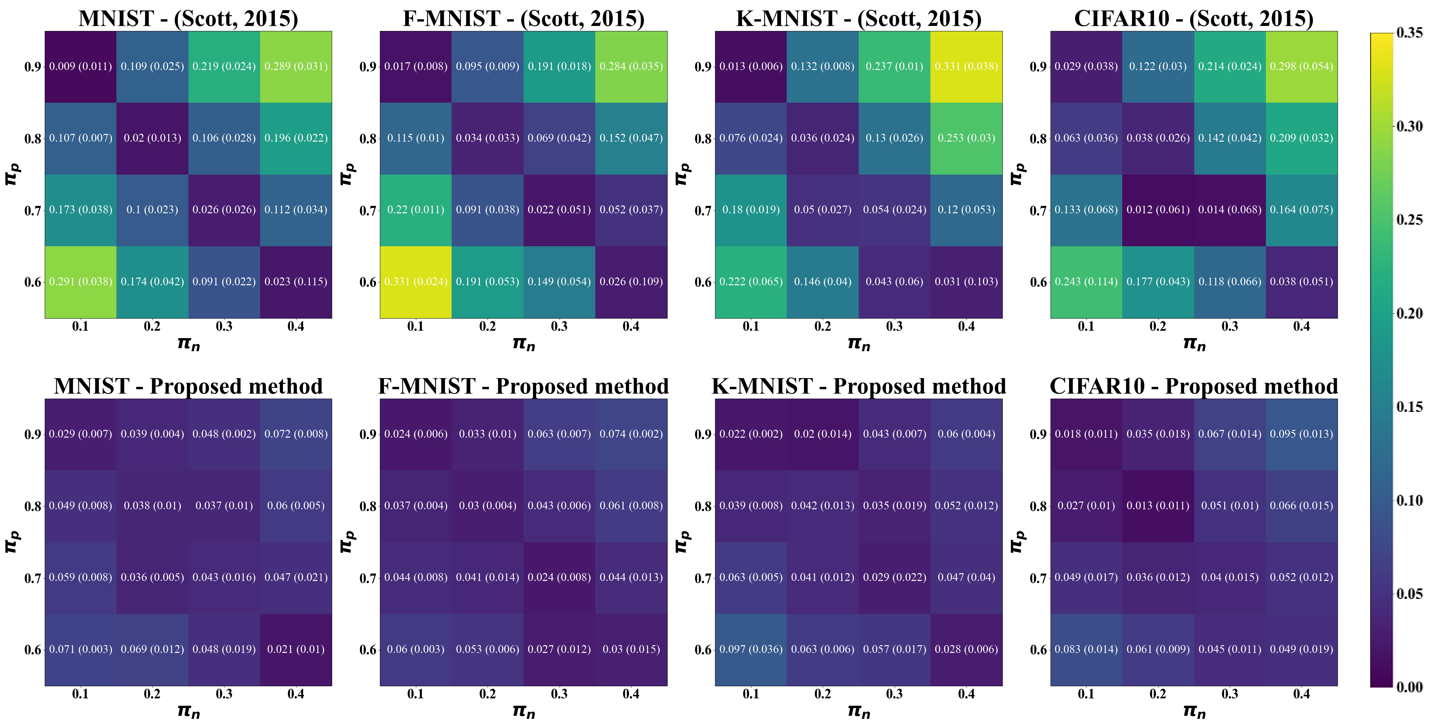}
    \begin{center}
  \caption{The heatmap of class-prior estimation errors for \cite{scott2015rate} (top) and our proposed method (bottom). Different color lumps show different extents of estimation errors when meeting various combinations of true $\pi_\mathrm{n}$ and $\pi_\mathrm{p}$. The numerical values on color lumps are the mean of estimation errors with stand deviations. We ran the estimation process five times.}
    \label{Fig:compare with scott}
    \end{center}
    \vspace{-2em}
\end{figure*}

Previously, \cite{scott2015rate} applied a mutual MPE model~\cite{scott2015rate} to the setting of learning from noisy labels with the rigorous assumption of class priors. Although their method relates to our work, its direct application in our MU-OPPO setting has drawbacks. Firstly, the algorithm in~\cite{scott2015rate} is based on the mutual MPE model. Hence, it is required to pair all unlabeled datasets first when meeting $m (m > 2)$ unlabeled datasets. The pairing should satisfy the Assumption \ref{ass:m_classprior} that implies that the number of positive samples in noisy (corrupted) positive distribution is larger than the number of negative samples in  noisy (corrupted) negative distribution. Although this assumption guarantees the subsequent estimation of class priors, it is hard to check and satisfy in our MU-OPPO setting. We know only one pairwise numerical relationship of class priors and cannot access any specific class priors. Secondly, in the final formula (Eq.~(\ref{invert_estimate})) of estimating class priors, there are estimates $\hat{\tilde{\pi}}_{\mathrm{n}}$ and $\hat{\tilde{\pi}}_{\mathrm{p}}$ in denominators when estimating class priors. If the numerical values of estimates $\hat{\tilde{\pi}}_{\mathrm{n}}$ and $\hat{\tilde{\pi}}_{\mathrm{p}}$ are large or small, the final estimation error would be high. Consequently, this creates an unstable problem when estimating $\pi_{\mathrm{n}}$ and $\pi_{\mathrm{p}}$, which is clearly shown in  Figure~\ref{Fig:compare with scott}. Next, we summarize the method of~\cite{scott2015rate}, then discuss how it differs from our proposal in terms of estimating class priors through experiments.

In~\cite{scott2015rate}, a mutual MPE model was proposed:
\begin{align} \label{scott1}
\tilde{\mathbbm{P}}_{\mathrm{p}}(\boldsymbol{x})=\pi_{\mathrm{p}} \mathbbm{P}_{\mathrm{p}}(\boldsymbol{x})+(1-\pi_{\mathrm{p}})\mathbbm{P}_{\mathrm{n}}(\boldsymbol{x}),\\
\label{scott2}
\tilde{\mathbbm{P}}_{\mathrm{n}}(\boldsymbol{x})=\pi_{\mathrm{n}} \mathbbm{P}_{\mathrm{p}}(\boldsymbol{x})+(1-\pi_{\mathrm{n}})\mathbbm{P}_{\mathrm{n}}(\boldsymbol{x}),
\end{align}
where $\pi_{\mathrm{p}}$ and $\pi_{\mathrm{n}}$ are the class priors of noisy (corrupted) positive and negative distribution $\tilde{\mathbbm{P}}_{\mathrm{p}}$ and $\tilde{\mathbbm{P}}_{\mathrm{n}}$, respectively. Note that \cite{scott2015rate} makes the strict assumption of class priors as follows.
\begin{assumption} \label{ass:m_classprior}
Assume that $\pi_{\mathrm{p}} > \frac{1}{2}$ \text{and} $\pi_{\mathrm{n}} < \frac{1}{2}$.
\end{assumption}
According to Proposition 3 in~\cite{scott2015rate}, Assumption \ref{ass:m_classprior} implies $\pi_{\mathrm{p}} > \pi_{\mathrm{n}}$. If $\mathbbm{P}_{\mathrm{p}} \neq \mathbbm{P}_{\mathrm{n}}$, then $\tilde{\mathbbm{P}}_{\mathrm{p}} \neq \tilde{\mathbbm{P}}_{\mathrm{n}}$, and there exists unique $0 \leq \tilde{\pi}_{\mathrm{p}}, \tilde{\pi}_{\mathrm{n}} < 1$ that could substitute above Eq.~(\ref{scott1}) and (\ref{scott2}) to:
\begin{align} 
\tilde{\mathbbm{P}}_{\mathrm{p}}(\boldsymbol{x})=(1-\tilde{\pi}_{\mathrm{p}}) \mathbbm{P}_{\mathrm{p}}(\boldsymbol{x})+\tilde{\pi}_{\mathrm{p}}\tilde{\mathbbm{P}}_{\mathrm{n}}(\boldsymbol{x}),\\
\tilde{\mathbbm{P}}_{\mathrm{n}}(\boldsymbol{x})=(1-\tilde{\pi}_{\mathrm{n}}){\mathbbm{P}}_{\mathrm{n}}(\boldsymbol{x})+\tilde{\pi}_{\mathrm{n}} \tilde{\mathbbm{P}}_{\mathrm{p}}(\boldsymbol{x}).
\end{align}
In particular, 
\begin{align}
\label{estimate}
\tilde{\pi}_{\mathrm{p}}=\frac{1-{\pi}_{\mathrm{p}}}{1-{\pi}_{\mathrm{n}}} \qquad \text{and} \qquad \tilde{\pi}_{\mathrm{n}}=\frac{{\pi}_{\mathrm{n}}}{{\pi}_{\mathrm{p}}}.
\end{align}
Furthermore, according to~\cite{scott2015rate}, if $\mathbbm{P}_{\mathrm{p}}$ and $\mathbbm{P}_{\mathrm{n}}$ satisfy the Assumption~\ref{ass:m_irr}, then the estimations $\hat{\tilde{\pi}}_{\mathrm{p}}$ and $\hat{\tilde{\pi}}_{\mathrm{n}}$ of $\tilde{\pi}_{\mathrm{p}}$ and $\tilde{\pi}_{\mathrm{n}}$ could be obtained by Eq.~(\ref{estimator1}), i.e, $\hat{\tilde{\pi}}_{\mathrm{p}} = \kappa^{*}(\tilde{\mathbbm{P}}_{\mathrm{p}}|\tilde{\mathbbm{P}}_{\mathrm{n}})$ and $\hat{\tilde{\pi}}_{\mathrm{n}} = \kappa^{*}(\tilde{\mathbbm{P}}_{\mathrm{n}}|\tilde{\mathbbm{P}}_{\mathrm{p}})$. We use these terms to estimate the class priors $\pi_{\mathrm{p}}$ and $\pi_{\mathrm{n}}$ by inverting the identities in Eq. (\ref{estimate}), leading to the estimation
\begin{align}
\label{invert_estimate}
\hat{\pi}_{\mathrm{p}}=\frac{1-\hat{\tilde{\pi}}_{\mathrm{p}}}{1-\hat{\tilde{\pi}}_{\mathrm{p}}\hat{\tilde{\pi}}_{\mathrm{n}}} \qquad \text{and} \qquad \hat{\pi}_{\mathrm{n}}=\frac{(1-\hat{\tilde{\pi}}_{\mathrm{p}})\hat{\tilde{\pi}}_{\mathrm{n}}}{1-\hat{\tilde{\pi}}_{\mathrm{p}}\hat{\tilde{\pi}}_{\mathrm{n}}}.
\end{align}

After obtaining $\hat{\pi}_{\mathrm{p}}$ and $\hat{\pi}_{\mathrm{n}}$, the cost parameter $\alpha \in (0,1)$ that is guaranteed by Assumption \ref{ass:m_classprior}, could be estimated by 
\begin{equation}
\hat{\alpha} = \frac{\hat{\pi}_{\mathrm{p}}-\frac{1}{2}}{\hat{\pi}_{\mathrm{p}}-\hat{\pi}_{\mathrm{n}}}.
\end{equation}
The cost parameter then is utilized to construct the \emph{$\alpha$-cost-sensitive $\tilde{P}$-risk}~\cite{scott2015rate}. For any binary classifier $f$, the $\alpha$-cost-sensitive $\tilde{P}$-risk is 
\begin{equation}
R_{\tilde{P},\hat{\alpha}}(f) \coloneqq \mathbb{E}_{(\bm{x},\tilde{y})\sim \tilde{P}}[(1-\hat{\alpha})\mathbbm{1}_{\{\tilde{y}=1\}}\mathbbm{1}_{\{f(\bm{x}) \leq 0\}} + \hat{\alpha}\mathbbm{1}_{\{\tilde{y}=0\}}\mathbbm{1}_{\{f(\bm{x}) > 0\}}],
\end{equation}
where $\tilde{P}$ denotes the probability measure governing $(\bm{x},\tilde{y})$. \cite{natarajan2013learning} show that minimizing a cost-sensitive $\tilde{P}$-risk is equivalent to minimizing the cost-insensitive $P$-risk:
\begin{equation}
R_{\mathrm{p}}(f) \coloneqq \mathbb{E}_{(\bm{x},y)\sim P}[\mathbbm{1}_{sign(f(\bm{x})) \neq y}],
\end{equation}
where $P$ denote the probability measure governing $(\bm{x},y)$. Therefore, the binary classifier $f$ could be obtained from $\tilde{\mathbbm{P}}_{\mathrm{p}}$ and $\tilde{\mathbbm{P}}_{\mathrm{n}}$ without the exact class priors $\pi_{\mathrm{p}}$ and $\pi_{\mathrm{n}}$. Note that in our setting, we also obtain $\tilde{\mathbbm{P}}_{\mathrm{p}}$ and $\tilde{\mathbbm{P}}_{\mathrm{n}}$ after assigning pseudo labels by one pairwise numerical relationship of class priors. Thus, we can empirically compare the method \cite{scott2015rate} and our proposal. For comparing the method \cite{scott2015rate} and our proposal, we conduct experiments in Figure~\ref{Fig:compare with scott}, which use a variety of unlabeled dataset pairs. In Figure~\ref{Fig:compare with scott}, the considerable variation of the colors in the top four plots reflects the unstable problem of \cite{scott2015rate}. By contrast, our method handles this problem well. 

\vspace{-10pt}

\section{Theoretical Justifications} \label{sec:5}

In our MOS framework for the proposed MU-OPPO setting, estimating class priors plays an essential role. Therefore, we conduct the theoretical analysis for estimation errors of class priors after confident example collection. Our solution framework is capable of incorporating various methods directly into different modules. Therefore, we set the confident example collection module based on latent representations and the class prior estimation module based on the standard MPE estimator. This is taken as an example of theoretical analysis. Here, we analyze the properties of confident example selection and theoretically justify its impact on estimating class priors. For simplicity, we omit the index and directly use $\mathbbm{P}_\mathrm{u}$ and $\pi$ to replace the $\mathbbm{P}_\mathrm{u}^j$ and $\pi_j$ in the following theoretical descriptions. 

\subsection{The Causes of Estimation Errors of Class Priors} \label{sec:5.1}
According to \cite{blanchard2010semi,scott2013classification,scott2015rate}, the standard MPE estimator is consistent and convergent to true class priors with strong theoretical guarantees. In our implementation, we primarily select the confident positive examples $\mathcal{X}_\mathrm{p}$ to approximate the distribution $\mathbbm{P}_\mathrm{p}$. However, the selection of $\mathcal{X}_\mathrm{p}$ may be corrupted by mixing the examples from the distribution $\mathbbm{P}_\mathrm{n}$. The selected examples  $\mathcal{X}_\mathrm{p}$ may be generated from $\mathbbm{P}_\mathrm{p}'$, where $\mathbbm{P}_\mathrm{p}'$ is a corrupted version of $\mathbbm{P}_\mathrm{p}$ and is obtained by mixing a small negative set from $\mathbbm{P}_\mathrm{n}$ to $\mathbbm{P}_\mathrm{p}$. Hence, the inaccurate performance of confident example selection is the main reason for estimation errors. In order to formally present the theoretical analysis, we  generate $\mathbbm{P}_{\mathrm{p}'}$ by splitting, transporting, and combining a set $E$ from $\mathbbm{P}_\mathrm{n}$ to $\mathbbm{P}_\mathrm{p}$. We detail the procedure as follows.

Let $M$ be a probability distribution on a measurable space $(\mathcal{X},\Omega)$. Given a set $E\in\Omega$, according to~\cite{yao2022rethinking}, it could be defined a distribution $M^E$ on the $\sigma$-algebra $\Omega$, i.e., $\forall S \in \Omega, M^E(S) = M(S\cap E)$. Therefore, given two distributions $M^E$ and $M^{E^c}$, where $E^c = \mathcal{X} \backslash E$, for any set $E\in \Omega$, we have $M^E + M^{E^c} = M$. Then, considering splitting the set $E \in \Omega$ from $\mathbbm{P}_\mathrm{n}$, $\mathbbm{P}_\mathrm{n}$ are divided into two parts: $\mathbbm{P}_\mathrm{n}^{E^c}$ and $\mathbbm{P}_\mathrm{n}^E$. That $\mathbbm{P}_\mathrm{n}^E$ is transported to $\mathbbm{P}_\mathrm{p}$, i.e., 
\begin{equation}\label{eq:divide_mix}
\mathbbm{P}_{\mathrm{u}} = \pi\mathbbm{P}_{\mathrm{p}} + (1-\pi)\mathbbm{P}_{\mathrm{n}}  = \pi \mathbbm{P}_{\mathrm{p}}+(1-\pi)\underbrace{(\mathbbm{P}_{\mathrm{n}}^E+\mathbbm{P}^{E^c}_{\mathrm{n}})}_{\text{divide into two}} = \underbrace{(\pi\mathbbm{P}_{\mathrm{p}} + (1-\pi)\mathbbm{P}_{\mathrm{n}}^E)}_{\text{mix as one}}+(1-\pi)\mathbbm{P}_{\mathrm{n}}^{E^c}.
\end{equation}
From Eq.~(\ref{eq:divide_mix}), due to the transport by $E$, we need to redefine $\mathbbm{P}_\mathrm{u}$ as a mixture of a corrupted distribution $\mathbbm{P}_{\mathrm{p}'}$ and a split distribution $\mathbbm{P}_{\mathrm{n}'}$ defined in Theorem~\ref{confusingtheorem}. 

\begin{theorem}[\cite{yao2022rethinking}] \label{confusingtheorem} 
Suppose $\mathbbm{P}_\mathrm{u} = \pi \mathbbm{P}_\mathrm{p} + (1-\pi)\mathbbm{P}_\mathrm{n}$ and $E \subset \text{Supp} (\mathbbm{P}_\mathrm{n})$. By splitting $\mathbbm{P}_\mathrm{n}^E$ from $\mathbbm{P}_\mathrm{n}$ to $\mathbbm{P}_\mathrm{p}$, $\mathbbm{P}_\mathrm{u}$ is a new mixture, i.e., $\mathbbm{P}_\mathrm{u} = \pi' \mathbbm{P}_{\mathrm{p}'} + (1-\pi')\mathbbm{P}_{\mathrm{n}'}$, where
\begin{equation}
    \pi'=\pi+(1-\pi)\mathbbm{P}_{\mathrm{n}}(E),~\mathbbm{P}_{\mathrm{n}'} = \frac{\mathbbm{P}_\mathrm{n}^{E^c}}{\mathbbm{P}_{\mathrm{n}}(E^c)},~\text{and}~\mathbbm{P}_{\mathrm{p}'}=\frac{(1-\pi){\mathbbm{P}_\mathrm{n}^E}+\pi \mathbbm{P}_\mathrm{p}}{(1-\pi)\mathbbm{P}_{\rm{n}}(E)+\pi},
\end{equation}
where $\mathbbm{P}_{\mathrm{p}'}$ and $\mathbbm{P}_{\mathrm{n}'}$ are satisfied the \textit{anchor point assumption}\footnote{The anchor point assumption \cite{Liu_2016,scott2015rate,yao2022rethinking} is a stronger variant of the irreducible assumption. It accelerates the convergence rate of a series of MPE estimators.}.
\end{theorem}
The proof of Theorem~\ref{confusingtheorem} can be found in \cite{yao2022rethinking}. According to Theorem \ref{confusingtheorem}, the new proportion $\pi'$ is always identifiable as $\mathbbm{P}_{\mathrm{p}'}$ and $\mathbbm{P}_{\mathrm{n}'}$ always satisfies the anchor set assumption. However, the $\pi'$ may not be close to $\pi$, which causes the estimation error. 

\begin{definition}[estimation error] \label{Estimation error}
Denote the estimation error by $\epsilon$ with $\epsilon = |\pi' - \pi| = (1-\pi)\mathbbm{P}_{\mathrm{n}}(E)$. 
\end{definition}
Note that, for $\epsilon$, the $\pi$ is a fixed latent value. Therefore, the item $\mathbbm{P}_{\mathrm{n}}(E)$ decides the extent of $\epsilon$.

\subsection{Theoretical Analysis of the Estimation Errors}
According to Definition~\ref{Estimation error}, the item $\mathbbm{P}_{\mathrm{n}}(E)$ decides the extent of $\epsilon$. Here, 
\begin{equation}
\mathbbm{P}_{\mathrm{n}}(E)=\sum_{\bm{x}_i\in E}\mathbbm{P}(\bm{x}_i|y_i=-1)
\end{equation}
reflects the negative class-conditional probability of the set $E$. In the procedure of confident example collection, this probability could be measured. For intuitive understanding, the set $E$ is made of the samples $(\bm{x}_i, y_i = -1)$ that are incorrectly selected as confident positive examples by our confident example collection module. Currently, our confident example collection module collects confident positive examples using latent representations, and we make the following reasonable assumptions referred to related works~\cite{kim2021fine,fisher1936use,lee2019robust}. Besides, we focus on the data points whose pseudo label is $\tilde{y}=+1$ in the following theoretical analysis. 

\begin{assumption}
The feature distribution is comprised of two Gaussian distributions. One is a clean cluster that contains the data points whose labels are $y = +1$ and $\tilde{y} = +1$. Another is an unclean cluster that   includes the data points whose labels are $y = -1$ and $\tilde{y} = +1$. The features of all instances with~$y =+1$ are aligned on the unit vector $\bm{v}$ with the white noise. Similarly, the features of all instances with~$y =-1$ are aligned on the unit vector $\bm{w}$.
\end{assumption}
As the feature distribution comprises two Gaussian distributions, the projected distribution $\bm{z}$ is also a mixture of two Gaussian distributions~\cite{kim2021fine}. By the linear discriminant
analysis (LDA) assumption \cite{fisher1936use,lee2019robust}, the decision boundary $B$ with the threshold $\zeta = 0.5 $ is the same as the average of the mean of two clusters. We have 
\begin{equation}\label{eq:boundary}
    B = \frac{1}{2}\left(\frac{\sum_{i=1}^{N}\mathbbm{1}_{\{\tilde{y}_{i} = +1,y_{i}=+1\}}\bm{z}_{i}}{N_{+}} + \frac{\sum_{i=1}^{N} \mathbbm{1}_{\{\tilde{y}_{i} = +1, y_{i}=-1\}} \bm{z}_{i}}{N_{-}}\right)
\end{equation}
with probability 1-$\delta$. In Eq.~(\ref{eq:boundary}), $N$ denotes the size of the set to be collected, i.e., the number of examples in a pair of unlabeled datasets. Also, $N^+=\sum_{i=1}^{N}\mathbbm{1}_{\{\tilde{y}_{i} = +1,y_{i}=+1\}}$ and $N^-=\sum_{i=1}^{N}\mathbbm{1}_{\{\tilde{y}_{i} = +1,y_{i}=-1\}}$. \cite{kim2021fine} proved that: 
\begin{equation}
\label{boundofb}
     \frac{(\bm{u}^{\top} \bm{v})^2 + (\bm{u}^{\top} \bm{w})^2}{2} - \mathcal{C} \sqrt{\frac{2}{N_{+}} \log(2/\delta)} \leq B \leq  \frac{(\bm{u}^{\top} \bm{v})^2 + (\bm{u}^{\top} \bm{w})^2}{2}+ \mathcal{C} \sqrt{\frac{2}{N_{+}}  \log(2/\delta)},
\end{equation}
where $\mathcal{C}>0$ is a constant. Then, based on Eq.~(\ref{boundofb}), we can derive the upper bound for the estimation error in the following Theorem~\ref{estimationerror}.

\begin{theorem}
\label{estimationerror}
Let $\Phi$ be the cumulative distribution function~(CDF) of $\mathcal{N}(0, 1)$. The corrupted set $E$ is selected as the wrong confident positive examples when the corresponding projection $\bm{z}_i>b$ and $\bm{z}_i \in E $. 
The upper bound of the estimation error $\epsilon$ can be derived as 
\begin{equation}
    \epsilon\leq |E| (1-\pi)\Phi\left(\frac{-\Delta + 2 \mathcal{C} \sqrt{\left(\frac{2}{N_{+}} \right) \log(2/\delta)}}{2 \sigma}\right),
\end{equation}
where  $\Delta = \left\Vert\bm{u}^{\top}\bm{v} - \bm{u}^{\top}\bm{w} \right\Vert_{2}^2$, $\pi$ is the class prior of an unlabeled dataset, $|E|$ denotes the sample size of $E$, $\sigma^2$ is a variance of white noise, $\delta$, and $\mathcal{C}$ are consistent numbers. 
\end{theorem}
The proof of Theorem~\ref{estimationerror} is shown in Appendix~\ref{sec:A}. Theorem~\ref{estimationerror} states that the upper bound for the estimation error $\epsilon$ depends on the latent class prior $\pi$ and $\Delta$ that denotes the difference of mean between two Gaussian distributions. A small upper bound can be guaranteed as class priors increase and $\Delta$ becomes larger. In this theoretical analysis, $\Delta$ is an essential factor related to the separation of \textit{alignment clusterability} (Definition~\ref{AlignmentClusterability}). Fortunately, this propriety has been ensured with the theoretical and empirical analysis by \cite{kim2021fine}. Therefore, in our solution framework, Theorem~\ref{estimationerror} guarantees the gap between estimated class priors and true class priors. As a result, with the estimated class priors, our MOS framework can obtain a well-worked binary classifier and tackles the proposed MU-OPPO setting.

\section{Experiments}\label{sec:6}

\begin{table}[h]
    \centering
    \begin{tabular}{l|cccc}
    \hline
         Datasets&\#~Train&\#~Test&\#~Features&$\pi_{\mathcal{D}}$ \\ \hline
         MNIST~\cite{lecun1998gradient}& 60,000&10,000&784&0.49\\
         Fashion (F)-MNIST~\cite{DBLP:journals/corr/abs-1708-07747}&60,000&10,000&784&0.80\\
         Kuzushiji (K)-MNIST~\cite{clanuwat2018deep}&60,000&10,000&784&0.30\\
         CIFAR-10~\cite{krizhevsky2009learning}&50,000&10,000&3,072&0.70\\
         
    \hline
    \end{tabular}
    \caption{The important statistics of used datasets.}
    \label{tab:datasets}
\end{table}

\paragraph{Datasets.} We verify the effectiveness of the proposed learning paradigm on widely adopted benchmarks, i.e., MNIST~\cite{lecun1998gradient}, Fashion (F)-MNIST~\cite{DBLP:journals/corr/abs-1708-07747}, Kuzushiji (K)-MNIST~\cite{clanuwat2018deep}, and CIFAR-10~\cite{krizhevsky2009learning}. The important statistics are shown in Table~\ref{tab:datasets}. As the four datasets contain ten classes originally, we manually corrupt them into binary-classification datasets as did in~\cite{lu2021binary}. In experiments, unless otherwise specified, the number
of examples contained in all unlabeled datasets is the same. In addition, the class priors $\{\pi_j\}_{j=1}^{m}$ of all unlabeled datasets are evenly generated from the range $[0.1, 0.9]$. The generation way makes the sampled class priors not all identical, which ensures that the problem is mathematically solvable~\cite{scott2020learning}.

\paragraph{Models \& Optimization.} We exploit ResNet-18~\cite{he2016deep} as our warm-up classifier, with the SGD optimizer~\cite{bottou2012stochastic}. The number of warm-up epochs is set to 10. After training the warm-up classifier, the models and optimizations of different modules in our MOS framework are consistent with their original paper. Note that, considering the computing efficiency, the pair-selection number $\gamma$ is set to 4 when employing the ECCPE to estimate class priors in our MOS framework.

\subsection{The Generalization of the MOS Framework}

\begin{table*}[h]\centering \small
    \newcommand{\tabincell}[2]{\begin{tabular}{@{}#1@{}}#2\end{tabular}}
        \begin{tabular}{ l | l| c |c |c |c } 
        \hline
        \tabincell{c}{Confident-Example \\ Collection} & \tabincell{c}{Class-Prior \\ estimation} & \tabincell{c}{MNIST} & \tabincell{c}{F-MNIST} & \tabincell{c}{K-MNIST} & \tabincell{c}{CIFAR-10}  \\ 
        \hline
        \multirowcell{3}{By the Example \\ Loss Distribution}
        &  Standard MPE & 3.07±0.71  &  4.76±0.88  & 3.99±1.18  & 3.78±0.92 \\
        \cline{2-6}

        & ReMPE & \textbf{2.73}±0.76 & \textbf{2.81}±0.73  & 4.08±1.37  &  3.44±0.86\\
        \cline{2-6}
        & BBE & 3.82±0.85    & 3.15±0.64  & 4.49±0.23   & 2.41±0.67  \\
		\hline
        \multirowcell{3}{By the Prediction \\ Probability}
        &  Standard MPE & 3.68±0.54  &  4.21±0.54  & 4.36±0.38  & 2.61±0.63 \\
        \cline{2-6}
        & ReMPE & 3.13±0.32 & 2.82±0.48  & 3.75±0.62  & 2.24±0.42 \\
        \cline{2-6}
        & BBE & 3.45±0.14    & 3.63±0.97  & 4.23±0.24   &  2.56±0.69\\
        		\hline
        \multirowcell{3}{By the Latent \\ Representations}
        &  Standard MPE & 3.50±0.26  & 3.73±0.50  & 3.92±0.75  & 3.17±0.55 \\
        \cline{2-6}
        & ReMPE & 2.89±0.26 & 2.82±0.58  & \textbf{3.28}±0.24  &   \textbf{2.15}±0.29\\
        \cline{2-6}
        & BBE & 3.65±0.24    & 2.97±1.15  & 4.75±0.17   & 2.72±0.37  \\
		\hline
        \end{tabular}
    \caption{Mean absolute estimation errors, multiplied by 100 with standard deviations of all class priors over five trials. We bold the smallest estimation errors by comparing different methods.}
    \label{estimation_error_generlization}
\end{table*} 

In MU-OPPO, our MOS framework is a generalized framework that could incorporate various methods in different modules. For the confident-example collection module, we apply three methods, i.e., with the loss distribution, confident learning, and latent representations, which have been introduced in Section~\ref{Confident-Example Collection module}. We apply three estimators for the class-prior estimation module, i.e., MPE, ReMPE, and BBE, introduced in Section~\ref{Class-Prior estimate module}. For binary classification with estimated class priors, we apply two methods, i.e., MCM and U$^m$-SSC, which have been introduced in Section~\ref{sec:3.4}. In this subsection, we experiment with all combinations of various methods in different modules. The binary classifier is learned from $m=10$ sets of unlabeled datasets with the assumed numerical relationship, e.g., $\pi_{10} > \pi_1$. We analyze the generalization of our proposed framework from the following two perspectives, i.e., the estimation error of class-prior estimating and the classification accuracy of the binary classifier. 

\begin{table*}[!t]\centering \small
    \newcommand{\tabincell}[2]{\begin{tabular}{@{}#1@{}}#2\end{tabular}}
        \begin{tabular}{ l | l| c |c |c |c} 
        \hline
        \tabincell{c}{Confident-Example \\ Collection} & \tabincell{c}{Class-Prior \\ estimation} & \tabincell{c}{MNIST} & \tabincell{c}{F-MNIST} & \tabincell{c}{K-MNIST} & \tabincell{c}{CIFAR-10}  \\ 
        \hline
        \multirowcell{3}{By the Example \\ Loss Distribution}
        &  Standard MPE & 96.60±0.25  &  93.51±0.28  & 89.38±0.67  &  86.24±0.27 \\
        \cline{2-6}
        & ReMPE & 96.41±0.41 & 93.69±0.21  & 89.20±0.56  &  86.20±0.23\\
        \cline{2-6}
        & BBE & 97.01±0.14    & \textbf{93.81}±0.12  & 90.23±0.38  & 86.17±0.22  \\

		\hline
        \multirowcell{3}{By the Prediction \\ Probability}
        &  Standard MPE & 96.83±0.25  &  93.52±0.20  & 90.08±0.26  & 86.10±0.15 \\
        \cline{2-6}
        & ReMPE & 96.81±0.26 & 93.78±0.17  & \textbf{90.31}±0.41  &  \textbf{86.26}±0.17 \\
        \cline{2-6}
        & BBE & 97.02±0.12    & 93.64±0.18  & 90.30±0.25   &  86.13±0.11\\
        \hline
        \multirowcell{3}{By the Latent \\ Representations}
        &  Standard MPE & 96.62±0.48  & 93.57±0.05  & 89.80±0.30  & 86.21±0.20 \\
        \cline{2-6}
        & ReMPE & 96.95±0.18 & 93.69±0.15  & 90.09±0.51  & 86.19±0.23 \\
        \cline{2-6}
        & BBE & \textbf{97.05}±0.16    & 93.74±0.12  & 90.27±0.29   & 86.22±0.13  \\
        \hline
	\multicolumn{2}{c|}{Given True Class Priors} & 97.09±0.15    & 93.89±0.08  & 90.35±0.39   &  86.29±0.13\\
        \hline

        \end{tabular}
    \caption{Means of the classification accuracy with standard deviations over five trials when doing binary classification by the MCM module. The methods that provide the highest accuracy are highlighted in boldface.}
    \label{binary_classifier_MCM}
\end{table*}

\paragraph{Estimation error of class-prior estimating.} In the proposed MOS framework, the main component is the estimation of class priors and we employ ECCPE here. This part aims to validate whether the proposed framework could reduce the estimation error of class priors and if this property could be preserved when applying different methods. To have a rigorous performance evaluation, each case runs five times and reports the mean and standard deviation. The evaluation is the average of absolute estimation errors between all class priors, shown in Table \ref{estimation_error_generlization}. We can see that the MOS framework could estimate class priors with a minor estimation error. Furthermore, we find that applying different methods to our framework also could perform reasonable estimations. This makes us believe that our framework not only provides accurate class-prior estimation but also shows the generalization capacity of various methods when estimating class priors. In addition, the ReMPE estimator provides significantly better performance on estimating the class priors in most cases since the regrouping process in the ReMPE estimator could alleviate the violation of the irreducible assumption in practice.

\begin{table*}[h]\centering \small
    \newcommand{\tabincell}[2]{\begin{tabular}{@{}#1@{}}#2\end{tabular}}
        \begin{tabular}{ l | l| c |c |c |c} 
        \hline
        \tabincell{c}{Confident-Example \\ Collection} & \tabincell{c}{Class-Prior \\ estimation} & \tabincell{c}{MNIST} & \tabincell{c}{F-MNIST} & \tabincell{c}{K-MNIST} & \tabincell{c}{CIFAR-10}  \\ 
        \hline
        \multirowcell{3}{By the Example \\ Loss Distribution}
        &  Standard MPE & \textbf{97.77}±0.11  &  94.07±0.15  & 91.20±0.29  & 87.01±0.12 \\
        \cline{2-6}
        & ReMPE & 97.75±0.10 & 94.18±0.10  & 91.18±0.12  &  \textbf{87.04}±0.08\\
        \cline{2-6}
        & BBE & 97.70±0.11    & 94.29±0.11  & 91.63±0.36  & 86.91±0.13 \\
	\hline
        \multirowcell{3}{By the Prediction \\ Probability}
        &  Standard MPE & 97.72±0.11  & 94.31±0.13  & 91.46±0.38  & 86.90±0.15 \\
        \cline{2-6}
        & ReMPE & 97.73±0.10 & 94.32±0.03  & 91.49±0.28  &  87.01±0.13\\
        \cline{2-6}
        & BBE & 97.76±0.07    & 94.25±0.11  & 91.62±0.21   & 86.83±0.12  \\
	\hline
        \multirowcell{3}{By the Latent \\ Representations}
        &  Standard MPE &  97.69±0.10  &  94.33±0.10  & \textbf{91.69}±0.38  & 86.94±0.15 \\
        \cline{2-6}
        & ReMPE & 97.70±0.07 & \textbf{94.33}±0.09  & 91.49±0.27  &  86.95±0.07\\
        \cline{2-6}
        & BBE & 97.72±0.04    & 94.27±0.14  & 91.61±0.26   &  86.97±0.10\\
		\hline
		\multicolumn{2}{c|}{Given True Class Priors} & 97.78±0.09    & 94.36±0.09  & 91.71±0.23   &  87.08±0.09\\
        \hline
        \end{tabular}
    \caption{Means of the classification accuracy with standard deviations over five trials when binary classification by the U$^m$-SSC module. The methods that provide the highest accuracy are highlighted in boldface.}
    \label{binary_classifier_Um}
\end{table*} 

\paragraph{Classification accuracy of binary classifiers.} In this part, we train a binary classifier from the MU-OPPO learning scheme using our MOS framework, which contains various methods of different modules. All experiments are trained by 300 epochs. The classification accuracy at the last epoch in the test phase is reported. All the experiments are repeated five times. The mean accuracy with standard deviations is recorded for each method. For clarity, we divide all the experiments into Table~\ref{binary_classifier_Um} and Table~\ref{binary_classifier_MCM} by two different binary classification modules, i.e., MCM and U$^m$-SSC. Checking the results in Table~\ref{binary_classifier_Um}, the classification accuracy of our proposed framework is highly close to the one given the true class priors in the MU-OPPO problem. The utilization of different methods within our framework also can perform well. This property also could be found in Table~\ref{binary_classifier_MCM}. Hence, we confirm that our proposed framework successfully solves the MU-OPPO problem, and the empirical performance is ideal. Besides, if we compare the experiment results from  Table~\ref{binary_classifier_MCM} and Table~\ref{binary_classifier_Um}, we could find that the experiment results in Table~\ref{binary_classifier_Um} are better than Table~\ref{binary_classifier_MCM}. The reason is that the optimal combination weights in the MCM method are proved with strong model assumptions and thus remain difficult to be tuned in practice~\cite{lu2021binary}. These results are consistent with the observations in~\cite{lu2021binary}.

Suppose we connect the experiment results from the estimation errors of class-prior estimating and the classification accuracy of the binary classifier. In that case, we also notice that some methods that have the smallest estimation error may not provide the highest classification accuracy, e.g., the framework utilizes the loss distribution as the confident-example collection module and use the ReMPE estimator as the class-prior estimation module has the lowest estimation error (2.73$\pm$0.76) but do not provide highest classification accuracy (97.75$\pm$0.10). This is because the U$^m$-SSC methods that classify with class priors are slightly robust to inaccurate class priors~\cite{lu2021binary}. 

\subsection{Comparison with Other Baselines} \label{Compare with some designed baselines}

\paragraph{Baselines.} Note that, in MU-OPPO, our MOS framework is the first solution that can estimate the class priors while other related works fail on this problem (described in Section~\ref{sec:2.2}). Therefore, we carefully design these baselines as follows:
\begin{itemize}
    \item MOS-M: Here, we want to introduce the method of \cite{scott2015rate} as our baseline, but this method is infeasible in our MU-OPPO setting, which has been clearly described in Section~\ref{sec:4}. Thus, we apply the mutual model of~\cite{scott2015rate} as a class priors estimator after pairing all unlabeled datasets and regarding it as the \underline{MOS}-\underline{m}utual~(MOS-M). The details are provided in Appendix \ref{mutualmodel}.
    \item MOS-$(\cdot)$: In the MOS process, the class prior estimation module also can be replaced by the other related class-prior estimators: the kernel mean embedding-based estimators KM1 and KM2~\cite{ramaswamy2016mixture}, a non-parametric class prior estimator AlphaMax~(AM)~\cite{jain2016nonparametric}, Elkan-Noto~(EN)~\cite{elkan2008learning}, DEDPUL~(DPL) \cite{ivanov2019dedpul}, and Rankprunning~(RP)~\cite{northcutt2017learning}. Then, we implement them under our framework and call them MOS-$(\cdot)$, which places the abbreviation name of these estimators in $(\cdot)$.
    \item MOS-E: The \underline{E}CCPE is employed to estimate class priors in the \underline{MOS} framework.    
    \item MOS-C: The \underline{C}CPE is employed to estimate class priors in the \underline{MOS} framework.
    \item MOS-T: To better show the performance of our proposed framework, we give the true class priors to the MU-OPPO setting and then implement \underline{MOS} framework with \underline{t}rue class priors.

\end{itemize}

Due to the generalization of our framework, we casually choose the method based on latent representations as the confident-example collection module and the classical MPE estimator as the class-prior estimation module in the MOS framework.  If not specified, we keep this setting in all subsequent experiments. We compare our proposed method with designed baseline methods for the MU-OPPO problem. The binary classifier is learned from $m=10$ sets of unlabeled datasets with the assumed numerical relationship, e.g., $\pi_{10} > \pi_1$.

\begin{table*}[h]\centering 
    \newcommand{\tabincell}[2]{\begin{tabular}{@{}#1@{}}#2\end{tabular}}
        \begin{tabular}{ c|c| c |c |c} 
        \hline
        Method & \tabincell{c}{MNIST} & \tabincell{c}{F-MNIST} & \tabincell{c}{K-MNIST} & \tabincell{c}{CIFAR-10}  \\ 
        \hline
        MOS-(KM1)    
        & 7.34$\pm$0.25  & 6.53$\pm$0.53  & 8.73$\pm$0.84  & 12.13$\pm$0.15 \\
		\hline
        MOS-(KM2)
        & 7.16$\pm$1.13  & 7.06$\pm$0.54  & 7.78$\pm$0.63  & 11.67$\pm$0.64 \\

		\hline
		MOS-(AM)      
        & 3.84$\pm$1.03    & 5.23$\pm$0.64  & 6.84$\pm$0.33   & 11.34$\pm$1.12  \\

		\hline
		MOS-(EN)      
        & 9.14$\pm$3.33  & 30.03$\pm$4.96  & 31.9$\pm$10.45  & 38.64$\pm$15.03 \\
        
        \hline
 		MOS-(DPL)      
        & 4.25$\pm$1.73  & 5.84$\pm$0.63  & 5.91$\pm$2.82  & 10.7$\pm$1.83  \\
        
        \hline
		MOS-(RP)    
        & 8.23$\pm$0.82  & 7.13$\pm$0.74   & 7.71$\pm$0.73  & 11.2$\pm$0.53 \\
                \hline
		MOS-M    
        & 4.12$\pm$0.83  & 5.84$\pm$2.13  & 5.63$\pm$0.64  & 7.43$\pm$0.93   \\
        \hline
		MOS-C~(ours)    
        & 4.23$\pm$0.32  & 5.93$\pm$1.21 & 4.21$\pm$0.82  &  3.92$\pm$1.14 \\
        
        \hline
		MOS-E~(ours)     
        & \textbf{3.50}$\pm$0.26  & \textbf{3.73}$\pm$0.50 & \textbf{3.92}$\pm$0.75 &\textbf{3.17}$\pm$0.55  \\
        
        \hline
        \end{tabular}
    \caption{Mean absolute estimation errors multiplied by 100 with standard deviations of all class priors over five trials. We bold the smallest estimation errors by comparing different methods.}
    \label{estimation_baseline}
    \end{table*} 

\paragraph{Class-prior estimation.} This part aims to validate whether the proposed method could reduce the estimation error of class priors compared to previous work. Each estimation case runs five times to have a rigorous performance evaluation and gets the mean and standard deviation. The evaluation is the average of absolute estimation errors between all class priors, shown in Table \ref{estimation_baseline}. We can see that MOS-E estimates class priors with a smaller estimation error compared to other baselines through all datasets. Furthermore, we find that our MOS-C method also performs reasonable estimation. It is better than MOS-M in the K-MNIST and CIFAR-10 datasets. 

\begin{table*}[h]\centering

    \newcommand{\tabincell}[2]{\begin{tabular}{@{}#1@{}}#2\end{tabular}}
        \begin{tabular}{ c| c | c |c |c} 
        \hline
        Method & \tabincell{c}{MNIST} & \tabincell{c}{F-MNIST} & \tabincell{c}{K-MNIST} & \tabincell{c}{CIFAR-10}  \\ 
        \hline
        MOS-(KM1)    
        & 97.53$\pm$00.21  & 93.74$\pm$0.14  & 91.06$\pm$0.37  & 86.22$\pm$0.08\\
		\hline
		MOS-(KM2)      
        & 97.57$\pm$00.29  & 93.49$\pm$0.14  & 91.21$\pm$0.03  & 86.19$\pm$0.10\\

		\hline
		MOS-(AM)    
        & 97.30$\pm$0.05  & 93.90$\pm$0.09  & 90.97$\pm$0.12  & 86.20$\pm$0.41 \\

		\hline
		MOS-(EN)   
        & 92.10$\pm$5.66  & 70.60$\pm$8.23  & 78.97$\pm$8.09   & 70.30$\pm$9.13 \\
        
        \hline
		MOS-(DPL)      
        & 97.63$\pm$0.06  & 93.76$\pm$0.08  & 91.16$\pm$0.21  & 86.15$\pm$0.10  \\
        
        \hline
		MOS-(RP)    
        & 96.63$\pm$0.02  & 93.57$\pm$0.07  & 91.02$\pm$0.06  & 85.92$\pm$0.44  \\
        
        \hline
		MOS-M   
        & 96.56$\pm$0.12  & 93.48$\pm$0.10  & 90.82$\pm$0.06  & 86.22$\pm$0.14 \\
        
        \hline
		MOS-C~(ours)    
        & 96.54$\pm$0.09  & 93.21$\pm$0.23  & 91.01$\pm$0.12  & 86.01$\pm$0.28  \\
        
        \hline
		MOS-E~(ours)    
        & \textbf{97.72}$\pm$0.11  & \textbf{94.31}$\pm$0.13 & \textbf{91.46}$\pm$0.38 &\textbf{86.90}$\pm$0.15  \\
        
        \hline
		MOS-T   
        & 97.78$\pm$0.09  & 94.36$\pm$0.09  & 91.71$\pm$0.23  & 87.08±0.09 \\
        
        \hline
        \end{tabular}
    \caption{Means of the classification accuracy with standard deviations over five trials. The methods that provide the highest accuracy are highlighted in boldface.}
    \label{classification_baseline}
\end{table*} 

\paragraph{Binary classification.} In this part, we train a binary classifier from the MU-OPPO learning scheme. All experiments are trained by 300 epochs. The classification accuracy at the last epoch in the test phase is reported in Table \ref{classification_baseline}. All the experiments are repeated five times. The mean accuracy with standard deviations is recorded for each method. Checking the results in Table \ref{classification_baseline}, our proposed MOS framework outperforms others in all cases. The classification accuracy is highly close to the MOS-T.

\subsection{On the Variation of the Set Number}

\begin{figure*}[!t]
    \centering
    \includegraphics[width=\linewidth]{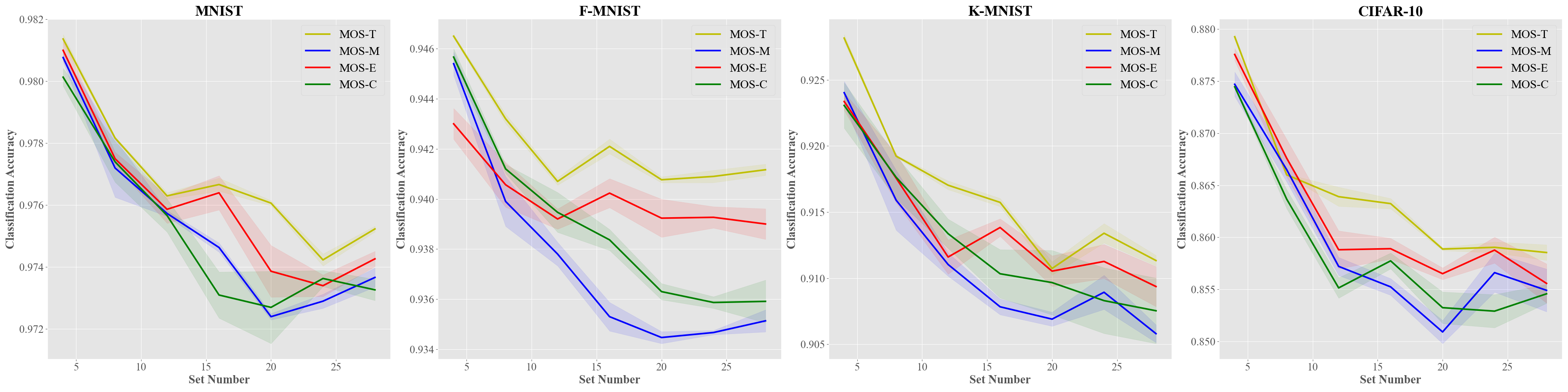}
    \begin{center}
  \caption{Classification accuracy with standard deviations for the proposed solution framework tested with different numbers of unlabeled datasets.}
    \label{Fig:Set_Number}
    \end{center}
    \vspace{-2em}
\end{figure*}

The main factor influencing the performance is the number of unlabeled datasets available. As the unlabeled sets can be easily collected from multiple sources~\cite{lu2021binary}, the learning algorithm is expected to perform well under the variation of set numbers. In this subsection, we test the proposed method with the MOS-M method on the different numbers of unlabeled datasets: $m = \{4,8,12,16,20,24,28\}$. We assume that the known relationship is $\pi_{m} > \pi_{1}$. We train 300 epochs for all the experiments, and the classification accuracy at the last epoch in the test phase is reported in Figure \ref{Fig:Set_Number}. From the results, we can see that the performance of the proposed MOS-E method is reasonably well on different set numbers. In most cases, the MOS-E method is better than the MOS-C and MOS-M. In particular, high accuracy can be observed for $m=4$ accessing all four benchmark datasets. The better performance may come from the larger number of unlabeled datasets contained in a single set, i.e., $n'/4$ in this case. The possible reason is that increasing the sampled data within each unlabeled set guarantees a better approximation of them~\cite{lu2021binary}. In addition, compared to MOS-C and MOS-M, the proposed MOS-E method demonstrates its effectiveness when the set numbers increase. This is a significant advantage, which shows the enhanced performance from MOS-C to MOS-E when the number of estimated class priors grows, and MOS-C is not steady in the large set number of cases (reflected by large shade areas).

\begin{table*}[h]
        \begin{center}
        \newcommand{\tabincell}[2]{\begin{tabular}{@{}#1@{}}#2\end{tabular}}
        \resizebox{\textwidth}{!}{
        \begin{tabular}{ l|l| c c c c | c} 
        \hline
        \tabincell{c}{Dataset} & \tabincell{c}{Method} &\tabincell{c}{$\tau=0.8$} & \tabincell{c}{$\tau=0.6$} & \tabincell{c}{$\tau=0.4$} & \tabincell{c}{\textbf{$\tau=0.2$}} & \tabincell{c}{{Random}}\\ 
        \hline
        \multirow{3}{*}{MNIST}
        & MOS-C & 97.54$\pm$0.11 & 97.58$\pm$0.04 & 97.59$\pm$0.04 &  97.46$\pm$0.10 & 97.37$\pm$0.20\\
		& MOS-E & \textbf{97.69}$\pm$0.03  & \textbf{97.64}$\pm$0.11 & \textbf{97.61}$\pm$0.13  &  \textbf{97.51}$\pm$0.23 & \textbf{97.84}$\pm$0.33\\
		 & MOS-T  & 97.71$\pm$0.06 & 97.66$\pm$0.04 &  97.63$\pm$0.26 &  97.79$\pm$0.02 &  97.88$\pm$0.26\\
		\hline
        \multirow{3}{*}{F-MNIST}      
        & MOS-C  & 94.10$\pm$0.08 & \textbf{94.27}$\pm$0.09 & \textbf{93.92}$\pm$0.17 & 91.61$\pm$0.54 &  93.15$\pm$0.63\\
		& MOS-E & \textbf{94.31}$\pm$0.06 & 94.06$\pm$0.06 &  93.81$\pm$0.09  & \textbf{92.13}$\pm$0.28 & \textbf{93.87}$\pm$0.20\\
		& MOS-T  & 94.34$\pm$0.15  & 94.10$\pm$0.22 & 94.20$\pm$0.16  & 93.32$\pm$0.38 & 94.04$\pm$0.30\\
		\hline
		\multirow{3}{*}{K-MNIST}      
        & MOS-C &92.22$\pm$0.43  & 91.78$\pm$0.41 & 91.71$\pm$0.31  & 91.03$\pm$0.23 & 90.41$\pm$0.40\\
		& MOS-E & \textbf{92.34}$\pm$0.42 & \textbf{91.96}$\pm$0.27 & \textbf{91.92}$\pm$0.16 & \textbf{91.11}$\pm$0.44 & \textbf{90.91}$\pm$0.39\\
		& MOS-T  & 92.39$\pm$0.35 & 92.07$\pm$0.63 & 92.33$\pm$0.30  &  91.24$\pm$0.42& 91.76$\pm$0.60\\
		\hline
		\multirow{3}{*}{CIFAR-10}      
        & MOS-C & 85.83$\pm$0.31 & 85.86$\pm$0.31 & 85.39$\pm$0.40 & 85.77$\pm$0.16 & 86.18$\pm$0.43\\
		& MOS-E & \textbf{85.98}$\pm$0.20 & \textbf{85.89}$\pm$0.39 & \textbf{86.05}$\pm$0.28 & \textbf{85.82}$\pm$0.43 & \textbf{86.20}$\pm$0.12\\
		& MOS-T  & 86.18$\pm$0.04  & 86.34$\pm$0.10 & 86.22$\pm$0.12 & 86.07$\pm$0.16 &  86.38$\pm$0.59\\
        \hline
        
        \end{tabular}}
        \end{center}
        \caption{Mean test accuracy with standard deviations over 3 trials tested on different set sizes. The uniform set size $n_j$ is shifted to $\tau\cdot n_j$ (smaller $\tau$, larger shift). ``Random'' means randomly sample a set size from the range $[0,n']$.  The methods that provide the closest accuracy to MOS-T are highlighted in boldface.}
        \label{size_shift_table}
\end{table*} 

\subsection{On the Variation of the Set Size}
In practice, the size of the unlabeled datasets may vary from an extensive range depending on different tasks, which may cause severe covariate shifts between training sets and test sets~\cite{lu2021binary,zhang2020one}. To verify the robustness of our proposed framework against set size shift, we conduct experiments on the variation of set sizes. Recall that in other experiments. We use uniform set sizes, i.e., all sets contain $n'/m$ unlabeled data. In this subsection, we investigate two set size shift settings according to \cite{lu2021binary}: (1) randomly select $\lceil m/2\rceil$ unlabeled datasets and change their set sizes to $\tau\cdot n'/m$, where $\tau\in [0, 1]$; (2) randomly sample each set size $n_j$ from range $[0,n']$ such that $\sum_{j = 1}^m n_j=n'$.

More specially, $m$ are set to 10. In the first design, we primarily generate unlabeled datasets like the setting in Section \ref{Compare with some designed baselines} and then change their set sizes. As shown in Table \ref{size_shift_table}, our proposed method is robust as $\tau$ moves towards 0 in the first shift setting. The accuracy degrades relatively slightly as $\tau$ decreases. We also observe that our solution framework performs well for four datasets in the second shift setting. In particular, MOS-E still has better robustness than MOS-C due to the lower standard deviation in many cases. Overall, the robustness of proposed methods on varied set sizes can be verified by changing set sizes. 

\subsection{Test Prior Estimation}

\begin{table}[!t]
\centering 
\newcommand{\tabincell}[2]{\begin{tabular}{@{}#1@{}}#2\end{tabular}}
\begin{center}
\begin{tabular}{ l| l |c c c c} 
\hline
Dataset & Method & \tabincell{c}{5 sets} & \tabincell{c}{10 sets} & \tabincell{c}{50 sets} & \tabincell{c}{100 sets}\\
\hline
\multirow{3}{*}{MNIST}
    & MOS-C  & 97.73$\pm$0.07 & 97.30$\pm$0.15 & 97.20$\pm$0.15 & 97.16$\pm$0.20\\
    & MOS-E  & \textbf{97.74}$\pm$0.06  & \textbf{97.45}$\pm$0.11  & \textbf{97.27}$\pm$0.15  & \textbf{97.30}$\pm$0.13 \\
    & MOS-T  &97.96$\pm$0.06   & 97.63$\pm$0.02  & 97.33$\pm$0.04  & 97.37$\pm$0.02 \\
\hline
\multirow{3}{*}{F-MNIST}
     & MOS-C  & 94.47$\pm$0.05 & 93.81$\pm$0.24 & 93.36$\pm$0.21 &  92.74$\pm$0.21\\
    & MOS-E  & \textbf{94.54}$\pm$0.13  & \textbf{93.82}$\pm$0.02  & \textbf{93.42}$\pm$0.46  & \textbf{92.78}$\pm$0.61 \\
    & MOS-T  & 94.65$\pm$0.01   & 94.12$\pm$0.03  & 94.06$\pm$0.02  & 93.83$\pm$0.06 \\
\hline
\multirow{3}{*}{K-MNIST}
     & MOS-C  & 90.61$\pm$0.28 &  90.29$\pm$0.76 &  89.32$\pm$0.42 &   \textbf{89.31}$\pm$0.42\\
    & MOS-E  & \textbf{91.34}$\pm$0.38  & \textbf{90.33}$\pm$0.19  & \textbf{89.99}$\pm$0.34  & 88.95$\pm$0.73 \\
    & MOS-T  &92.05$\pm$0.03   & 91.70$\pm$0.07  & 90.80$\pm$0.03  & 90.98$\pm$0.09 \\
\hline
\multirow{3}{*}{CIFAR-10}
     & MOS-C  & 87.06$\pm$0.12 &  85.21$\pm$0.11  & 84.52$\pm$0.45  &  83.76$\pm$0.74\\
    & MOS-E  & \textbf{87.07}$\pm$0.11  & \textbf{85.32}$\pm$0.37  & \textbf{84.62}$\pm$0.46  & \textbf{83.87}$\pm$0.21\\
    & MOS-T  &87.11$\pm$0.15   & 85.72$\pm$0.18  & 85.32$\pm$0.04  & 85.02$\pm$0.10 \\
\hline
\end{tabular}
\end{center}
\vspace{-1.5em}
\caption{Mean test accuracy with standard deviations over three trials when estimating both tests prior $\pi_{\mathcal{D}}$ and class priors $\pi_j$. The methods that provide the closest accuracy to MOS-T are highlighted in boldface.}
\label{test_prior_table}
\end{table}

Note that in MU-OPPO, the test prior $\pi_{\mathcal{D}}$, which was previously assumed known, could be estimated by our MOS framework. In this subsection, we learn a binary classifier from MU-OPPO after estimating both class priors $\pi_j$ and the test prior $\pi_{\mathcal{D}}$ by our method. We also verify the steady performance of proposed methods from small set numbers, e.g., $m=5$, to large set numbers, e.g., $m=100$. Three hundred epochs still perform all experiments. The assumed numerical relationship is $\pi_{m} > \pi_{1}$. Following the results in Table \ref{test_prior_table}, our observations are as follows. First, the proposed methods could still perform well when adding the estimation of the test prior  $\pi_{\mathcal{D}}$. Second, our proposed framework empirically guarantees the robustness of larger set numbers with acceptable standard deviations.

\subsection{Ablation Study}

\begin{table*}[h]
\footnotesize
 \centering

    \newcommand{\tabincell}[2]{\begin{tabular}{@{}#1@{}}#2\end{tabular}}
        \begin{tabular}{ l | c | c |c |c} 
        \hline
        Methods & \tabincell{c}{MNIST} & \tabincell{c}{F-MNIST} & \tabincell{c}{K-MNIST} & \tabincell{c}{CIFAR-10}  \\ 
        \hline
        MOS   
        & \textbf{97.72}$\pm$0.11  & \textbf{94.31}$\pm$0.13 & \textbf{91.46}$\pm$0.38 &\textbf{86.90}$\pm$0.15\\
        \hline
        MOS without Class Prior Estimation      
        & 94.80±0.59  & 91.46±0.71 &  87.86±0.63 & 82.11±0.36\\
        \hline
		MOS without Confident Example Collection   
        & 96.94±0.19  & 93.85±0.22 &  89.13±0.23 & 83.76±0.29\\
        \hline
		MOS without the Warm-Up Classifier      
        & 97.12±0.12  & 93.95±0.25  & 90.12±0.32  & 85.06±0.19\\

        \hline
        \end{tabular}
    \caption{Experimental results for ablation study. The methods that provide the highest accuracy are highlighted in boldface.}
    \label{Ablation_study}
\end{table*} 

We study the effect of removing different components to provide insights into what makes the MOS framework successful. We analyze the results in Table~\ref{Ablation_study} as follows.

\begin{itemize}

\item \textbf{MOS without Class-Prior estimation.} When we identify confident examples after our confident-example collection module, we could directly train a binary classifier on them. However, the confident examples we collected are hard to contain clean \textit{hard examples}~\cite{bai2021me}, which are essential to training a binary classifier. This is because the clean hard examples are usually entangled with mislabeled samples~\cite{bai2021me}. Thus, the trained classifiers are not optimal. To study the effect of this point, we directly use the collected confident examples and a ResNet-18 model to train a binary classifier. We report the classification accuracy at the last epoch in the test phase. The poor classification accuracy clearly reflects that the confident examples are not supposed to learn a classifier directly, but they can be used to learn the class priors.

\item \textbf{MOS without Confident Example Collection.} To study the effect of confident example collection, we drop the confident example collection module and directly estimate class priors by inputting the unlabeled datasets into the class-prior estimator. Note that the other training process remains unchanged. The performance decreases compared to the MOS framework and shows the necessity of collecting confident examples.

\item \textbf{MOS without the Warm-Up Classifier.} The reason for training a warm-up classifier is that early-stopping the training of deep learning models will prevent them from generally adapting to mislabeled data as training epochs become large. In this manner, a warm-up classifier whose training is stopped in early epochs will provide reliable guidance in selecting confident examples. Therefore, to  study the effect of the warm-up classifier, we replace it with a fully-trained (convergent) classifier that is trained with large training epochs. The other modules remain unchanged. The decrease in accuracy suggests that the warm-up classifier helps the selection of confident examples and indirectly benefits the training of the final binary classifier.

\end{itemize}

\section{Related Work}\label{sec:7}
We review more related literature on this work below. Remarkably, existing methods designed for binary classification from unlabeled datasets are varied from learning paradigms. 

Partial previous methods are regarded as \textit{discriminative clustering} based on \textit{maximum likelihood estimation}, such as maximizing the margin or the mutual information between given instances and unknown instance labels~\cite{mann2007simple,felix2013psvm,rueping2010svm}. Recall that clustering methods are often suboptimal since they require that one cluster exactly belongs to one class~\cite{chapelle2002cluster}, which is rarely satisfied in practice. Beyond clustering, \cite{du2013clustering} and \cite{menon2015learning} evidenced the possibility
of \textit{empirical balanced risk minimization} (EBRM) when learning from two unlabeled datasets. Both adopt the \textit{balanced error}, which is a special case of the classification error \cite{brodersen2010balanced} as the performance measure. Though EBRM methods do not need the
knowledge of class priors~\cite{charoenphakdee2019symmetric}, they assume the class prior is strictly balanced and only handles the case of two unlabeled datasets.

The MU-OPPO problem setting is also related to \textit{learning with label proportions} (LLP), with
a subtle difference in the unlabeled dataset generation\footnote{In most LLP works, the generation of unlabeled datasets relies on uniform sampling and may result in the same label proportion for all unlabeled datasets, which make the LLP problem computationally intractable~\cite{scott2020learning}.}. In LLP, each unlabeled dataset is associated with a proportion
label of different classes. The challenge in LLP is to train models using the weak supervision of proportion labels. To overcome this issue, \cite{https://doi.org/10.48550/arxiv.1402.5902} exploited a deep learning algorithm on LLP by introducing the \textit{empirical proportion risk minimization} (EPRM). Recently, \cite{DBLP:journals/corr/abs-1910-13188} combined EPRM with consistency regularization to obtain the state-of-the-art performance on LLP. However, EPRM is inferior to \textit{empirical risk minimization} (ERM) since its learning is not consistent.

A breakthrough in binary classification from unlabeled datasets is the proposal of ERM-based methods. Specifically, \cite{ICLR:Lu+etal:2019} proposed the unlabeled-unlabeled~(UU) classification that assumed $m=2$ and $\pi_1 > \pi_2$, and provided a risk-consistent UU method that constructs an equivalent expression of the classification risk. Then, it is shown in \cite{lu2020mitigating} that the UU method can take negative values which causes overfitting in the empirical training risk. Hence, they proposed a novel \textit{consistent risk correction} technique that is robust against overfitting and improves the UU method. Although these ERM-based methods are advantageous in
terms of flexibility and theoretical guarantees, they are limited to two unlabeled datasets. Therefore, \cite{lu2021binary} extended these previous methods for the general unlabeled dataset setting ($m\geq2$) by adding a transition layer and the proposed method is \textit{classifier-consistent}~\cite{patrini2017making}. In addition, \cite{scott2020learning} solved this general setting by firstly pairing all unlabeled datasets relying on class priors and then combining the risk estimator of each pair. Although existing methods successfully learn a binary classifier from multiple unlabeled datasets with theoretical guarantees, they heavily rely on the precise class priors, which has led to a non-trivial dilemma---the uncertainty of class priors can undesirably prevent the learning process. Yet, it is still unexplored how to learn a binary classifier from multiple
unlabeled datasets without the exact class priors. To the best of our knowledge, this work is the first attempt to get rid of precise class priors when learning from multiple unlabeled datasets. 

\section{Conclusion}\label{sec:8}

In this work, we propose a new learning setting, i.e., MU-OPPO, and construct the MOS framework that can both estimates class priors accurately and achieves high binary classification accuracy. Specifically, this framework is based on the class prior estimation, which estimates multiple class priors using only one pairwise numerical relationship of class priors. After that, the injection of estimated class priors subsequently achieves a statistically-consistent classifier. By establishing an estimation error bound, we also prove that these estimated class priors are close to the true class priors under some conditions. Extensive experiments demonstrate that the proposed MOS framework can successfully train binary classifiers from multiple unlabeled datasets, and is competitive with existing methods.

\newpage
\bibliographystyle{plainnat}
\bibliography{main.bib}

\newpage
\appendix
\newpage
\appendix
\section*{Appendix}
\section{Proofs of Estimation Errors}
\label{sec:A}
\begin{proof}
We use the similar proof skills of Theorem 2 of \cite{kim2021fine}. Specifically, let $|E|$ denote the sample size of $E$, we have 

\begin{equation}
        \begin{split}
            \mathbbm{P}_\mathrm{n}(E) &= \sum_{\bm{x}_i \in E} \mathbbm{P}(\bm{x}_i |y_i = -1) \approx |E| \mathbbm{P}(\bm{z} > B |y = -1)\\ &\leq |E| \mathbbm{P}(\bm{z} > \frac{(\bm{u}^{\top} \bm{v})^2 + (\bm{u}^{\top} \bm{w})^2}{2} - \mathcal{C} \sqrt{ \frac{2}{N_{+}} \log(2/\delta)} ~\bigg|~ y= -1)\\
            &= |E| \mathbbm{P}(\bm{z} > \frac{\mu_{+} + \mu_{-}}{2} - \mathcal{C} \sqrt{ \frac{2}{N_{+}}\log(2/\delta)} ~\bigg|~ y= -1)\\
            &= |E| \mathbbm{P}( \frac{\bm{z} - \mu_-}{\sigma} > \frac{\mu_{+} - \mu_{-}}{2\sigma} - \frac{\mathcal{C} \sqrt{\frac{2}{N_{+}}  \log(2/\delta)}}{\sigma})\\
            &= |E| \mathbbm{P}(\mathcal{N}(0, 1) > \frac{\Delta - 2 \mathcal{C} \sqrt{ \frac{2}{N_{+}}  \log(2/\delta)}}{2 \sigma}) \\
            &= |E| (1 - \mathbbm{P}(\mathcal{N}(0, 1) \leq \frac{\Delta - 2 \mathcal{C} \sqrt{ \frac{2}{N_{+}} \log(2/\delta)}}{2 \sigma})) \\
            &= |E| (1 - \Phi(\frac{\Delta - 2 \mathcal{C} \sqrt{\frac{2}{N_{+}}\log(2/\delta)}}{2 \sigma})) \\
            &= |E| \Phi(\frac{-\Delta + 2 \mathcal{C} \sqrt{\frac{2}{N_{+}} \log(2/\delta)}}{2 \sigma})
        \end{split}
    \end{equation}
Thus, we have the upper bound of the estimation error:
\begin{equation}
        \begin{split}
            \epsilon = (1-\pi)\mathbbm{P}_\mathrm{n}(E) \leq  (1-\pi)|E|\Phi(\frac{-\Delta + 2 \mathcal{C} \sqrt{\frac{2}{N_{+}}  \log(2/\delta)}}{2 \sigma})
        \end{split}
    \end{equation}

where  $\Delta = \left\Vert\bm{u}^{\top}\bm{v} - \bm{u}^{\top}\bm{w} \right\Vert_{2}^2$, $\pi$ is the class prior of an unlabeled dataset, $|E|$ denotes the sample size of $E$, $\sigma^2$ is a variance of white noise, $\delta$, and $\mathcal{C}$ are consistent numbers. 
\end{proof}

\section{A Mutual Model of Class Priors Estimation}
\label{mutualmodel}
After running CCPE first and obtaining the initialization of the class priors $\{\hat{\pi}_j\}_{j=1}^m$, we could re-index the unlabeled datasets by $t \in \{1, ..., \tbinom{m}{2}\}$. Let $(\mathcal{X}_{\rm{u}}^{t,+},\pi_{t}^{+})$ and $(\mathcal{X}_{\rm{u}}^{t,-},\pi_{t}^{-})$ constitute the $t$-th pair of bags, such that $\pi_{t}^{-} \leq \pi_{t}^{+}$. Then we could apply the mutual MPE model in Proposition 3 of \cite{scott2015rate} on all the pairs:

\begin{align}
\mathbbm{P}_{\rm{u}}^{t,+}(\boldsymbol{x})=\pi_{t}^{+} \mathbbm{P}_{\rm{p}}(\boldsymbol{x})+(1-\pi_{t}^{+})\mathbbm{P}_{\rm{n}}(\boldsymbol{x}),\\
\mathbbm{P}_{\rm{u}}^{t,-}(\boldsymbol{x})=\pi_{t}^{-} \mathbbm{P}_{\rm{p}}(\boldsymbol{x})+(1-\pi_{t}^{-})\mathbbm{P}_{\rm{n}}(\boldsymbol{x}). 
\end{align}

According to pseudo label assignment, $\mathbbm{P}_{\rm{u}}^{t,+}(\boldsymbol{x})$ and $\mathbbm{P}_{\rm{u}}^{t,-}(\boldsymbol{x})$ are seemed as the noisy positive and negative density $\tilde{\mathbbm{P}}_\mathrm{p}$ and $\tilde{\mathbbm{P}}_\mathrm{n}$:

\begin{align} 
\tilde{\mathbbm{P}}_{\rm{p}}(\boldsymbol{x})=\pi_{t}^{+} \mathbbm{P}_{\rm{p}}(\boldsymbol{x})+(1-\pi_{t}^{+})\mathbbm{P}_{\rm{n}}(\boldsymbol{x}),\\
\label{MPE equation 2}
\tilde{\mathbbm{P}}_{\rm{n}}(\boldsymbol{x})=\pi_{t}^{-} \mathbbm{P}_{\rm{p}}(\boldsymbol{x})+(1-\pi_{t}^{-})\mathbbm{P}_{\rm{n}}(\boldsymbol{x}).
\end{align}

Substituting the above equations, we have 

\begin{align} 
\tilde{\mathbbm{P}}_{\rm{p}}(\boldsymbol{x})=(1-\tilde{\pi}_{t}^{+}) \mathbbm{P}_{\rm{p}}(\boldsymbol{x})+\tilde{\pi}_{t}^{+}\tilde{\mathbbm{P}}_{\rm{n}}(\boldsymbol{x}),\\
\tilde{\mathbbm{P}}_{\rm{n}}(\boldsymbol{x})=(1-\tilde{\pi}_{t}^{-}){\mathbbm{P}}_{\rm{n}}(\boldsymbol{x})+\tilde{\pi}_{t}^{-} \tilde{\mathbbm{P}}_{\rm{p}}(\boldsymbol{x}),
\end{align}

where
\begin{align}
\label{change}
\tilde{\pi}_{t}^{+}=\frac{1-{\pi}_{t}^{+}}{1-{\pi}_{t}^{-}} \qquad \text{and} \qquad \tilde{\pi}_{t}^{-}=\frac{{\pi}_{t}^{-}}{{\pi}_{t}^{+}}
\end{align}
are two class priors. Then, we can obtain estimations $\hat{\tilde{\pi}}_{t}^{+}$ and $\hat{\tilde{\pi}}_{t}^{-}$ of $\tilde{\pi}_{t}^{+}$ and $\tilde{\pi}_{t}^{-}$ by $\kappa^{*}(\tilde{\mathbbm{P}}_\mathrm{p}|\tilde{\mathbbm{P}}_\mathrm{n})$ and $\kappa^{*}(\tilde{\mathbbm{P}}_\mathrm{n}|\tilde{\mathbbm{P}}_\mathrm{p})$. We use these terms to estimate the class priors $\pi_{t}^{+}$ and $\pi_{t}^{-}$ by inverting the identities in Eq. (\ref{change}), leading to the estimations

\begin{align}
\hat{\pi}_{t}^{+}=\frac{1-\hat{\tilde{\pi}}_{t}^{+}}{1-\hat{\tilde{\pi}}_{t}^{+}\hat{\tilde{\pi}}_{t}^{-}} \qquad \text{and} \qquad \hat{\pi}_{t}^{-}=\frac{(1-\hat{\tilde{\pi}}_{t}^{+})\hat{\tilde{\pi}}_{t}^{-}}{1-\hat{\tilde{\pi}}_{t}^{+}\hat{\tilde{\pi}}_{t}^{-}}.
\end{align}
\end{document}